\documentclass[]{article}

\usepackage{url}
\usepackage{appendix}
\usepackage{amsmath}
\usepackage{authblk}
\usepackage{graphicx}
\usepackage[margin=1in]{geometry}

\begin{document}

\title{Data-driven prediction of a multi-scale Lorenz 96 chaotic system using deep learning methods: Reservoir computing, ANN, and RNN-LSTM}


\author[1]{Ashesh Chattopadhyay}
\author[1,2]{Pedram Hassanzadeh}
\author[3,4]{Devika Subramanian}

\affil[1]{Department of Mechanical Engineering, Rice University}
\affil[2]{Department of Earth Environmental and Planetary Sciences, Rice University}
\affil[3]{Department of Electrical and Computer Engineering, Rice University}
\affil[4]{Department of Computer Science, Rice University}




\maketitle

\begin{abstract}
In this paper, the performance of three deep learning methods for predicting short-term evolution and for reproducing the long-term statistics of a multi-scale spatio-temporal Lorenz 96 system is examined. The methods are: echo state network (a type of reservoir computing, RC-ESN), deep feed-forward artificial neural network (ANN), and recurrent neural network with long short-term memory (RNN-LSTM). This Lorenz 96 system has three tiers of nonlinearly interacting variables representing slow/large-scale ($X$), intermediate ($Y$), and fast/small-scale ($Z$) processes. For training or testing, only $X$ is available; $Y$ and $Z$ are never known or used. We show that RC-ESN substantially outperforms ANN and RNN-LSTM for short-term prediction, e.g., accurately forecasting the chaotic trajectories for hundreds of numerical solver's time steps, equivalent to several Lyapunov timescales. The RNN-LSTM and ANN show some prediction skills as well; RNN-LSTM bests ANN. Furthermore, even after losing the trajectory, data predicted by RC-ESN and RNN-LSTM have probability density functions (PDFs) that closely match the true PDF, even at the tails. The PDF of the data predicted using ANN, however, deviates from the true PDF. Implications, caveats, and applications to data-driven and data-assisted surrogate modeling of complex nonlinear dynamical systems such as weather/climate are discussed.
\end{abstract}


\section{Introduction}  

Various components of the Earth system involve multi-scale, multi-physics processes and high-dimensional chaotic dynamics. These processes are often modeled using sets of strongly-coupled nonlinear partial differential equations (PDEs), which are solved numerically on supercomputers. As we aim to simulate such systems with increasing levels of fidelity, we need to increase the numerical resolutions and/or incorporate more physical processes from a wide range of spatio-temporal scales into the models. For example, in atmospheric modeling for predicting the weather and climate systems, we need to account for the nonlinear interactions across the scales of cloud microphysics processes, gravity waves, convection, baroclinic waves, synoptic eddies, and large-scale circulation, just to name a few, not to mention the fast/slow processes involved in feedbacks from the ocean, land, and cryosphere \cite{collins2006formulation,collins2011development,flato2011earth,bauer2015quiet,jeevanjee2017perspective}.

Solving the coupled systems of PDEs representing such multi-scale processes is computationally prohibitive for many practical applications. As a result, to make the simulations tractable, a few strategies have been developed, which mainly involve only solving for slow/large-scale variables and accounting for the fast/small-scale processes through surrogate models. In weather/climate models, for example, semi-empirical physics-based parameterizations are often used to represent the effects of processes such as gravity waves and moist convection in the atmosphere or sub-mesoscale eddies in the ocean \cite{stevens2013climate,hourdin2017art,garcia2017modification,jeevanjee2017perspective,schneider2017climate}. A more advanced approach is ``super-parameterization'', which for example, involves solving the PDEs of moist convection on a high-resolution grid inside each grid point of large-scale atmospheric circulation, whose governing PDEs (the Navier-Stokes equations) are solved on a coarse grid \cite{khairoutdinov2001cloud}. While computationally more expensive, super-parameterized climate models have been shown to outperform parameterized models in simulating some aspects of climate variability and extremes \cite{benedict2009structure,andersen2012moist,kooperman2018rainfall}. 

More recently, ``inexact computing'' has received attention from the weather/climate community. This approach involves reducing the computational cost of each simulation by decreasing the \emph{precision} of some of the less-critical calculations \cite{palem2014inexactness,palmer2014climate}, thus allowing the saved resources to be \emph{reinvested}, for example, in more simulations (e.g., for probabilistic predictions) and/or higher resolutions for critical processes \cite{duben2014use,duben2014benchmark,thornes2017use,hatfield2018improving}. One type of inexact computing involves solving the PDEs of some of the processes with single- or half-precision arithmetic, which requires less computing power and memory compared to the conventional double-precision calculations \cite{palem2014inexactness,duben2015opportunities,leyffer2016doing}.

In the past few years, the rapid algorithmic advances in artificial intelligence (AI), and in particular data-driven modeling, have been explored for improving the simulation and prediction of nonlinear dynamical systems \cite{schneider2017earth,kutz2017deep,gentine2018could,moosavi2018machine,wu2019enforcing,toms2019deep,brunton2019data,duraisamy2019turbulence,reichstein2019deep,lim2019predicting,scher2019generalization}. One appeal of the data-driven approach is that fast/accurate data-driven surrogate models that are trained on data from high-fidelity, computationally demanding simulations can be used to accelerate and/or improve the prediction and simulation of complex dynamical systems. Furthermore, for some poorly understood processes for which observational data are available (e.g., clouds), data-driven surrogate models built using such data might potentially outperform physics-based surrogate models \cite{schneider2017earth,reichstein2019deep}. Recent studies have shown promising results in using AI to build data-driven parameterizations for modeling of some atmospheric and oceanic processes \cite{rasp2018deep,brenowitz2018prognostic,gagne2019machine,o2018using,bolton2019applications,dueben2018challenges,watson2019applying,salehipour2019deep}. In the turbulence and dynamical systems communities, similarly encouraging outcomes have been reported                                                           \cite{ling2016reynolds,mcdermott2017ensemble,pathak2018model,rudy2018deep,vlachas2018data,mohan2019compressed,wu2019enforcing,raissi2019physics,zhu2019physics,mcdermott2019deep} .   
     
\emph{The objective of the current study is to make progress toward addressing the following question: Which AI-based data-driven technique(s) can best predict the spatio-temporal evolution of a multi-scale chaotic system, when only the slow/large-scale variables are known (during training) and are of interest (to predict during testing)?} We emphasize that unlike many other studies, our focus is not on reproducing long-term statistics of the underlying dynamical system (although that will be examined too), but on predicting the short-term  trajectory from a given initial condition. Furthermore, we emphasize that, following the problem formulation introduced by \cite{dueben2018challenges}, the system's state-vector is only partially known and the fast/small-scale variables are unknown even during training.

\begin{figure}[t]
\centering
\includegraphics[width=\textwidth]{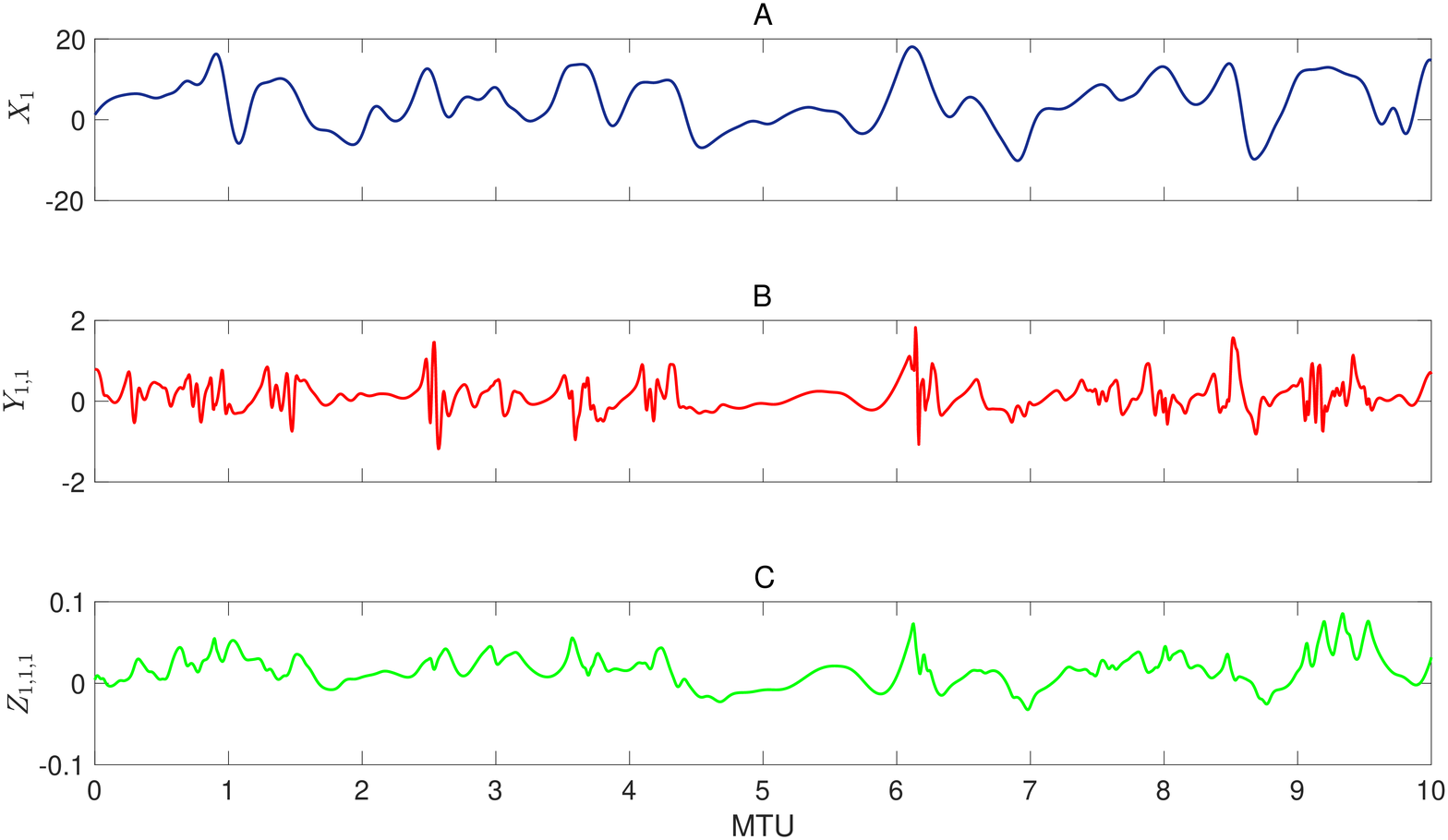}
\caption{Time evolution of (A) $X_k$ at grid point $k=1$,  (B) $Y_{j,1}$ at grid point $j=1$, and (C) $Z_{i,1,1}$ at grid point $i=1$. The time series show chaotic behavior in $X$, which has large amplitude and slow variability; $Y$, which has relatively small amplitudes, high frequency variability, and intermittency; $Z$, which has small amplitudes and high frequency variability. The $x$-axis is in model time unit (MTU) which is related to the time step of the numerical solver $(\Delta t)$ and largest positive Lyapunov exponent $(\lambda_{max})$ as $1$ MTU $=$ $200 \Delta t \approx 4.5/\lambda_{max}$ (see section~\ref{sec:RK4}). Note that here we are presenting raw data, which have not been standardized for prediction/testing yet.}
\label{xyz_demo}
\end{figure}

Our objective is more clearly demonstrated using the canonical chaotic system that we will use as a testbed for the data-driven methods: A multi-scale Lorenz 96 system:
\begin{eqnarray}
\label{lorenz}
\frac{dX_k}{dt}=X_{k-1}\left(X_{k+1}-X_{k-2}\right)+F-\frac{hc}{b}\Sigma_jY_{j,k} \label{eq:L1} \\ 
\frac{dY_{j,k}}{dt}=-cbY_{j+1,k}\left(Y_{j+2,k}-Y_{j-1,k}\right)-cY_{j,k}+  \frac{hc}{b}X_k -\frac{he}{d}\Sigma_iZ_{i,j,k} \label{eq:L2}\\
\frac{dZ_{i,j,k}}{dt}=edZ_{i-1,j,k}\left(Z_{i+1,j,k}-Z_{i-2,j,k}\right)- geZ_{i,j,k}+  \frac{he}{d}Y_{j,k} \label{eq:L3}
\end{eqnarray}
This set of coupled nonlinear ODEs is a 3-tier extension of Lorenz's original model \cite{lorenz1996predictability} and has been proposed by Thornes \textit{et al}.~ \cite{thornes2017use} as a fitting prototype for multi-scale chaotic variability of the weather/climate system and a useful testbed for novel methods. In these equations, $F=20$ is a large-scale forcing that makes the system highly chaotic; $b=c=e=d=g=10$ and $h=1$ are tuned to produce appropriate spatio-temporal variability in the three variables (see below). The indices $i, j, k = {1,2, \dots 8}$, thus $X$ has $8$ elements while $Y$ and $Z$ have $64$ and $512$ elements, respectively. Figure~\ref{xyz_demo} shows examples of the chaotic temporal evolution of $X$, $Y$, and $Z$ obtained from directly solving  Eqs.~(\ref{eq:L1})-(\ref{eq:L3}). The examples demonstrate that $X$ has large amplitudes and slow variability; $Y$ has relatively small amplitudes, high-frequency variability, and intermittency; and $Z$ has small amplitudes and high-frequency variability. In the context of atmospheric circulation, the slow variable $X$ can represent the low-frequency variability of the large-scale circulation while the intermediate variable $Y$ and fast variable $Z$ can represent synoptic/baroclinic eddies and small-scale convection, respectively. Similar analogies in the context of ocean circulation and various other natural or engineering systems can be found, making this multi-scale Lorenz 96 system a useful prototype to focus on. 

\begin{figure}[t]
\centering
\includegraphics[width=\textwidth]{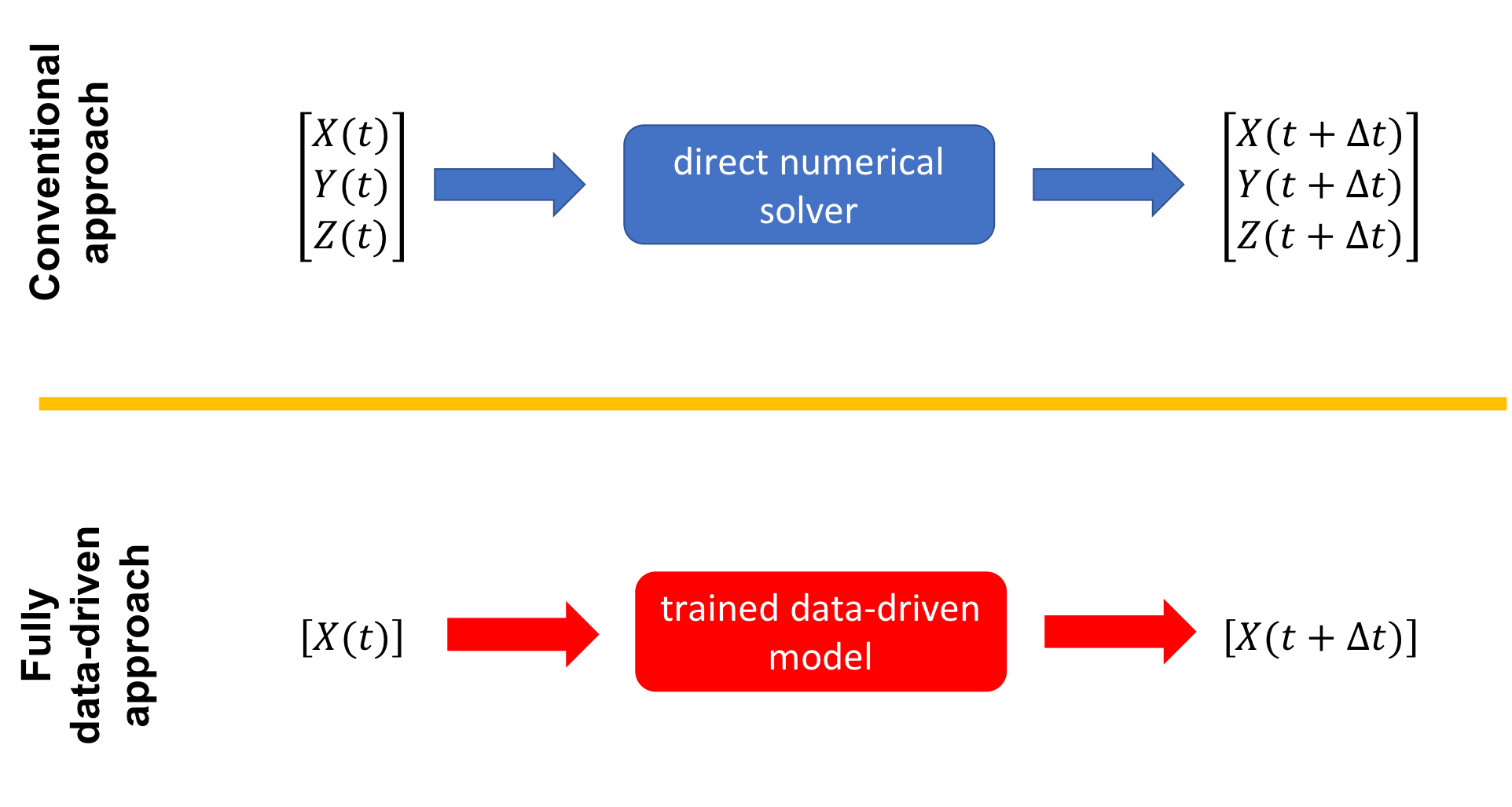}
\caption{The conventional approach of solving the full governing equations numerically versus our data-driven approach for predicting the spatio-temporal evolution of the slow/large-scale variable $X$. For the direct numerical solution, the governing equations have to be known and the numerical resolution is dictated by the fast and small-scale variable $Z$, resulting in high computational costs. In the data-driven approach, the governing equations do not have to be known, $Y$ and $Z$ do not have to be known, and evolution of $X$ is predicted just from knowing the past observations of $X$, leading to low computational costs.}
\label{data_drive_demo}
\end{figure}

\emph{Our objective is to predict the spatio-temporal evolution of $X(t)$ using a data-driven model that is trained on past observations of $X(t)$} (Figure~\ref{data_drive_demo}). In the conventional approach of solving Eqs.~(\ref{eq:L1})-(\ref{eq:L3}) numerically, the governing equations have to be known, initial conditions for $Y(t)$ and $Z(t)$ have to be available, and the numerical resolution is dictated by the fast/small-scale variable $Z$, leading to high computational costs. In the fully data-driven approach that is our objective here, the governing equations do not have to be known, $Y(t)$ and $Z(t)$ do not have to be observed at any time, and evolution of $X(t)$ is predicted just from knowledge of the past observations of $X$, leading to low computational costs. To successfully achieve this objective, a data-driven method should be able to
\begin{enumerate}
\item Accurately predict the evolution of a chaotic system along a trajectory for some time, 
\item Account for the effects of $Y(t)$ and $Z(t)$ on the evolution of $X(t)$. 
\end{enumerate}

Inspired by several recent studies (that are discussed below), we have focused on evaluating the performance of three deep learning techniques in accomplishing (1) and (2). These data-driven methods are
\begin{itemize}
\item RC-ESN: Echo state network (ESN), a specialized type of recurrent neural network (RNN), which belongs to the family of reservoir computing (RC),  
\item ANN: A deep feed-forward artificial neural network, 
\item RNN-LSTM: An RNN with long short-term memory (LSTM).   
\end{itemize}

We have focused on these three methods because they have either shown promising performance in past studies (RC-ESN and ANN), or they are considered state of the art in learning from sequential data (RNN-LSTM). There is a growing number of studies focused on using deep learning techniques for data-driven modeling of chaotic and turbulent systems, for example to improve weather/climate modeling and prediction. Some of these studies have been referenced above. Below, we briefly describe three sets of studies with closest relevance to our objective and approach. Pathak and co-workers \cite{pathak2018model,pathak2017using,lu2017reservoir} have recently shown promising results with RC-ESN for predicting short-term spatio-temporal evolution (item~1) and in replicating attractors for the Lorenz 63 and {Kuramoto-Sivashinsky} equations. The objective of our paper (items 1-2) and the employed multi-scale Lorenz~96 system are identical to that of \cite{dueben2018challenges}, who reported their ANN to have some skills in data-driven prediction of the spatio-temporal evolution of $X$. Finally, Vlachas \textit{et al}.~\cite{vlachas2018data} found an RNN-LSTM to have skills (though limited) for predicting the short-term spatio-temporal evolution (item~1) of a number of chaotic toy models such as the original Lorenz~96 system.      

Here we aim to build on these pioneering studies and examine, \textit{side by side}, the performance of RC-ESN, ANN, and RNN-LSTM in achieving (1) and (2) for the chaotic multi-scale Lorenz 96 system. {We emphasize the need for such a side-by-side comparison that uses the exact same system as the testbed and metrics for assessing performance.} 

The structure of the paper is as follows: in section~\ref{sec_method}, the multi-scale Lorenz 96 system and the three deep learning methods are discussed; results on how these methods predict the short-term spatio-temporal evolution of $X$ and reproduce the long-term statistics of $X$ are presented in section~\ref{sec_results}; key findings and future work are discussed in section~\ref{sec:discussion}.


\section{Materials and Methods}
\label{sec_method}
\subsection{The multi-scale Lorenz 96 system}\label{sec:RK4}
\subsubsection{Numerical solution}
We have used a 4th-order Runge-Kutta solver with time step $\Delta t=0.005$ to solve the system of Eqs.~(\ref{eq:L1})-(\ref{eq:L3}). The system has been integrated for $100$ million time steps to generate a large dataset for training, testing, and examining the robustness of the results.  
In the Results section, we often report time in terms of model time units (MTUs), where 1 MTU $= 200\Delta t$. In terms of Lyapunov timescale, 1 MTU in this system is $\approx 4.5/\lambda_{max}$ \cite{thornes2017use}, where $\lambda_{max}$ is the largest positive Lyapunov exponent. In terms of the $e$-folding decorrelation timescale ($\tau$) of the leading principal component (PC1) of $X$ \cite{khodkar2019reduced}, we estimate 1 MTU $\approx 6.9 \tau_\mathrm{PC1}$. Finally, as discussed in Thornes \textit{et al}.~ \cite{thornes2017use}, 1 MTU in this system corresponds to $\approx 3.75$ Earth days in the real atmosphere. 

\subsubsection{Training and testing datasets}
To build the training and testing datasets from the numerical solution, we have sampled $X \in \Re^{8}$ at every $\Delta t$ and then standardized the data by subtracting the mean and dividing by the standard deviation. We then construct a training set consisting of $N$ sequential samples and a testing consisting of the next $2000$ sequential samples from point $N+1$ to $N+2000$. We have randomly chosen $100$ such training/testing sets, each of length $(N+2000)\Delta t$ starting from a random point and separated from the next training/testing set by at least $2000 \Delta t$. There is never any overlap between the training and the testing sequences in any of the $100$ training/testing sets.
 
\subsection{Reservoir computing-echo state network (RC-ESN)}
\subsubsection{Architecture}
\begin{figure}[t]
\centering
\includegraphics[width=\textwidth]{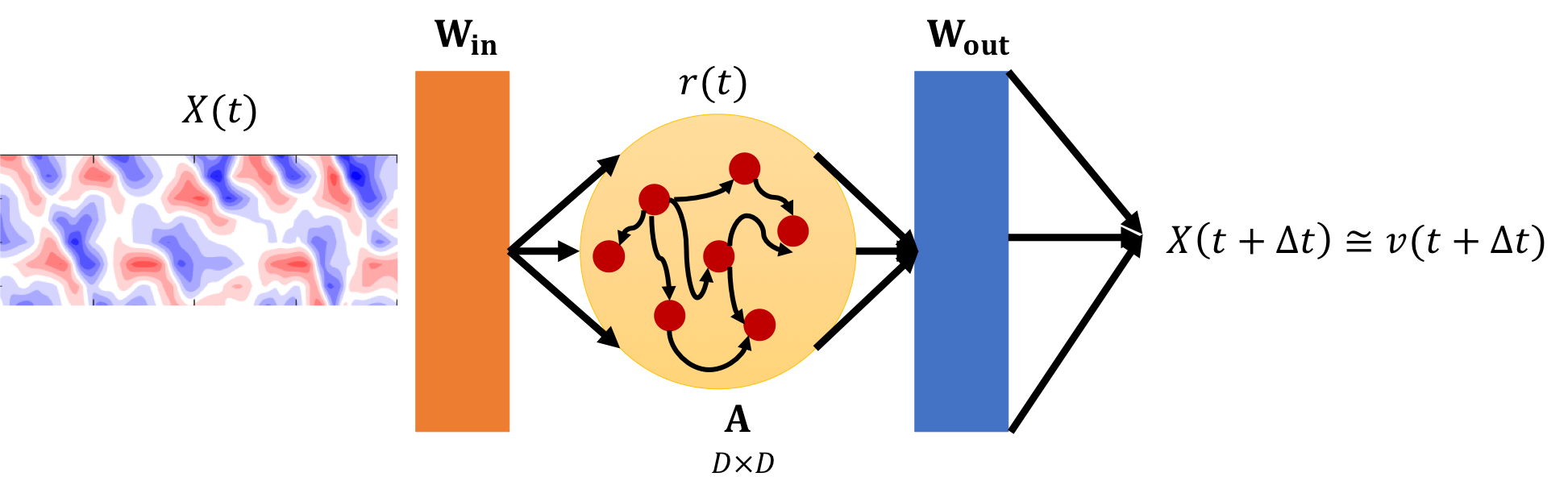}
\caption{A schematic of RC-ESN. $D \times D$ is the size of the adjacency matrix of the reservoir $\mathbf{A}$. $X(t)$ ($-T \leq t \leq 0 $) is the input during training. $\mathbf{W_{in}}$ and $\mathbf{A}$ are random matrices that are chosen at the beginning of training and fixed during training and testing. Only $\mathbf{W_{out}}$  is computed during training. During testing ($t>0$), $X(t+\Delta t)$ is predicted from a given $X(t)$ that is either known from the initial condition or has been previously predicted.}
\label{esn_demo}
\end{figure}

The RC-ESN \cite{jaeger2004harnessing,Jaeger2007} is an RNN that has a reservoir with $D$ neurons, which are sparsely connected in an Erd\H{o}s-R\'enyi graph configuration (see Figure~\ref{esn_demo}). The connectivity of the reservoir neurons is represented by the adjacency matrix $\mathbf{A}$ of size $D\times D$ whose values are drawn from a uniform random distribution on the interval $\left[-1 ,1 \right]$. The state of the reservoir, representing the activations of its constituent neurons, is a vector $r(t) \in \Re^{D}$.  The typical reservoir size used in this study is $D=5000$ (but we have experimented with $D$ as large as $20000$, as discussed later). The other two components of the RC-ESN are an input-to-reservoir layer with weight matrix $\mathbf{W_{in}}$,  and a reservoir-to-output layer with weight matrix $\mathbf{W_{out}}$. The inputs for training are $N$ sequential samples of $X(t) \in \Re^{8}$. At the beginning of the training phase, $\mathbf{A}$ and $\mathbf{W_{in}}$ are chosen randomly and are fixed; i.e., they do not change during training or testing. During training, only the weights of the output-to-reservoir layer ($\mathbf{W_{out}}$) are updated. 

$\mathbf{W_{out}}$ is the only trainable matrix in this network. This architecture yields a simple training process that has two key advantages:
\begin{itemize}
\item It does not suffer from the vanishing and the exploding gradient problem, which has been a major difficulty in training RNNs, especially before the advent of LSTMs \cite{pascanu2013difficulty},
\item $\mathbf{W_{out}}$ can be computed in one shot (see below), thus this algorithm is orders of magnitude faster than the backpropagation through time (BPTT) algorithm 
\cite{goodfellow2016deep}, which is used for training general RNNs and RNN-LSTMs (see section~\ref{LSTM}). 
\end{itemize}

The equations governing the RC-ESN training process are as follows:	
\begin{eqnarray}
\label{esn_eqn}
r(t+\Delta t)=\tanh\left(\mathbf{A}r(t)+\mathbf{W_{in}}X(t)\right), \label{eq:r} \\
\mathbf{W_{out}}=\arg\min_{\mathbf{W_{out}}} ||{\mathbf{W_{out}}\tilde{r}(t)-X(t)}||+\alpha ||\mathbf{W_{out}}||. \label{eq:wout} 
\end{eqnarray}
Here $|| \cdot ||$ is the $L_2$-norm of a vector and $\alpha$ is the $L_2$ regularization (ridge regression) constant. Equation~(\ref{eq:r}) maps the observable $X(t) \in \Re^{8}$ to the higher dimensional reservoir's state $r(t+\Delta t) \in \Re^{D}$ (reminder: $D$ is $O(1000)$ for this problem). Note that Eq.~(\ref{eq:wout}) contains $\tilde{r}(t)$ rather than $r(t)$. As observed by Pathak \textit{et al}.~ \cite{pathak2018model} for the {Kuramoto-Sivashinsky} equation, and observed by us in this work, the columns of the matrix $\tilde{\mathbf{r}}$ should be chosen as nonlinear combinations of the columns of the reservoir state matrix ${\mathbf{r}}$ (each row of the matrix $\mathbf{r}$ is a snapshot $r(t)$ of size $D$). For example, following Pathank \textit{et al}.~ \cite{pathak2018model}, we compute $\tilde{\mathbf{r}}$ as having the same even columns as ${\mathbf{r}}$ while its odd columns are the square of the odd columns of ${\mathbf{r}}$ (algorithm $T_1$ hereafter). As shown in Appendix~\ref{sec:appnT}, we have found that a nonlinear transformation (between ${\mathbf{r}}$ and $\tilde{\mathbf{r}}$) is essential for skillful predictions, while several other transformation algorithms yield similar results as $T_1$. The choices of these transformations ($T_2$ and $T_3$, see Appendix~\ref{sec:appnT}), although not based on a rigorous mathematical analysis, are inspired from the nature of the quadratic nonlinearity that is present in Eqs.~(\ref{eq:L1})-(\ref{eq:L3}). 

The prediction process is governed by:
\begin{eqnarray}
\label{esn_eqn_test}
v(t+\Delta t)=\mathbf{W_{out}}\tilde{r}(t+\Delta t), \label{eq:v} \\
X(t+\Delta t)=v(t+\Delta t). \label{eq:tdt}
\end{eqnarray}
$\tilde{r}(t+\Delta t)$ in Eq.~(\ref{eq:v}) is computed by applying one of the $T_1$, $T_2$ or $T_3$ algorithms on $r(t+\Delta t)$, which itself is calculated via Eq.~(\ref{eq:r}) from $X(t)$ that is either known from initial condition or has been previously predicted.

See \cite{jaeger2004harnessing,Jaeger2007,lukovsevivcius2009reservoir,gauthier2018reservoir} and references therein for further discussions on RC-ESNs, and \cite{lu2017reservoir,pathak2017using,mcdermott2017ensemble,pathak2018model,pathak2018hybrid,zimmermann2018observing,lu2018attractor,mcdermott2019deep,lim2019predicting} for examples of recent applications to dynamical systems. 

\subsubsection{Training and Prediction}
\label{train_esn}
During training ($-T \leq t \leq 0 $), $\mathbf{W_{in}}$ and $\mathbf{A}$ are initialized with random numbers, which stay fixed during the training (and testing) process, The state matrix $\mathbf{r}$ is initialized to $0$. During initialization, we ensure that the spectral radius of $\mathbf{A}$ is less than unity by first dividing the matrix $\mathbf{A}$ with its largest eigenvalue $\Lambda$ and further multiplying it by a scalar ($\rho \leq 1$). Then the state matrix $\mathbf{r}$ is computed using Eq.~(\ref{eq:r}) for the training set, and $\mathbf{W_{out}}$ is computed using Eq.~(\ref{eq:wout}). During prediction (i.e., testing) corresponding to $t > 0$, the computed $\mathbf{W_{out}}$ is used to march $v(t)$ (and thus $X(t)$) forward in time (Eqs.~(\ref{eq:v})-(\ref{eq:tdt})) while as mentioned earlier, $r(t)$ keeps getting updated using the predicted $X(t)$ (Eq.~(\ref{eq:r})). A non-linear transformation is used to compute $\tilde{r}(t)$ from $r(t)$ before using Eq.~(\ref{eq:v}). 

Our RC-ESN architecture and training/prediction procedure are similar to the ones used in \cite{pathak2018model}. There is no overfitting in the training phase because the final training and testing accuracies are the same. Our code is developed in Python and is made publicly available (see Code and data availability).

\subsection{Feed-forward artificial neural network (ANN)}
\label{section_ann}
\subsubsection{Architecture}
\begin{figure}[t]
\centering
\includegraphics[width=\textwidth]{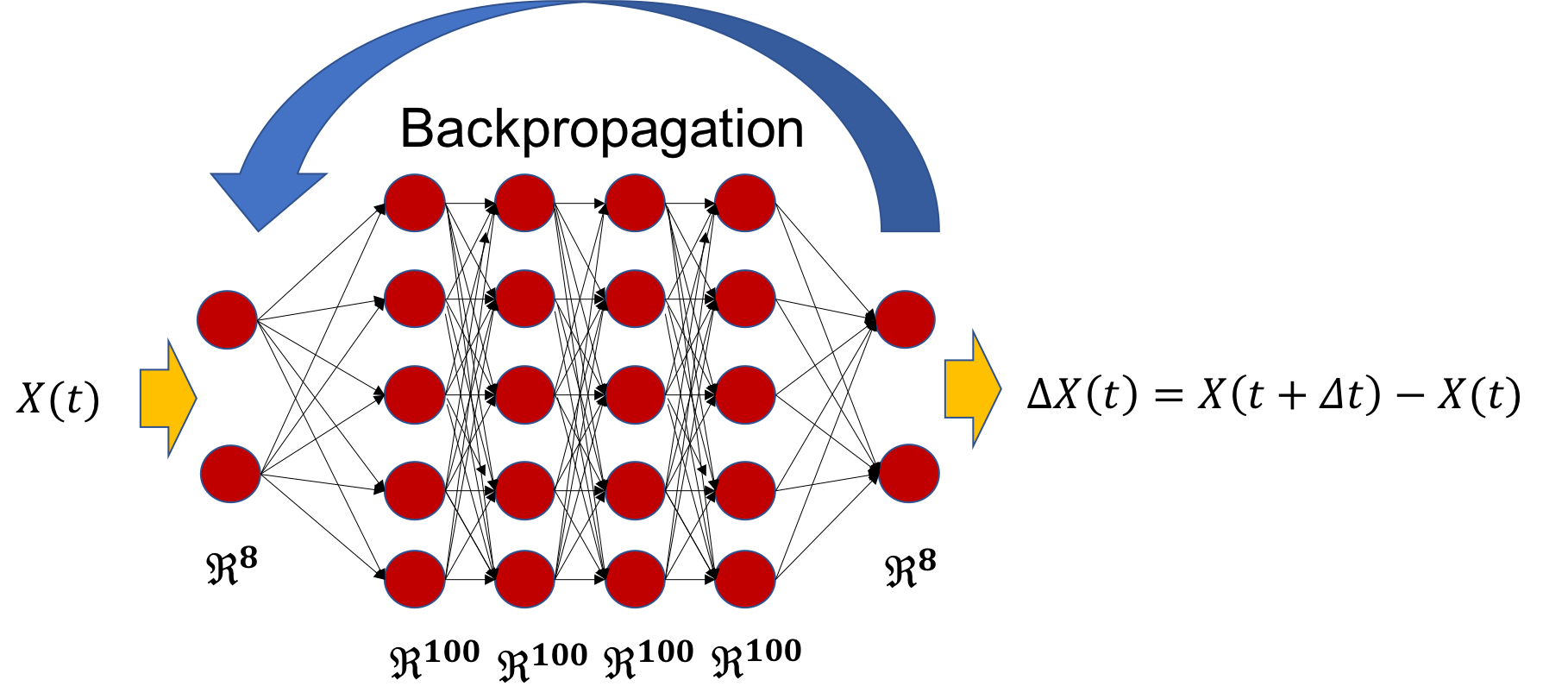}
\caption{A schematic of ANN, which has $4$ hidden layers each with $100$ neurons, and an input and an output layer each with $8$ neurons (for better illustration only few neurons from each layer are shown). Training is performed on pairs of $X(t)$ and $\Delta X(t)=X(t+\Delta t)-X(t)$, and the weights in each layer are learned through the backpropagation algorithm. During testing (i.e., prediction) $\Delta X(t)$ is predicted for a given $X(t)$ that is already predicted or is known from initial condition. Knowing $X(t)$, $\Delta X(t)$, and $\Delta X(t-\Delta t)$, $X(t+\Delta t)$ is then computed using Adams-Bashforth integration scheme.}
\label{ann_demo}
\end{figure}

We have developed a deep ANN that has the same architecture as the one used in Duben and Bauer~\cite{dueben2018challenges} (which they employed to conduct prediction on the same multi-scale Lorenz 96 studied here). The ANN has $4$ hidden layers with $100$ neurons each, and $8$ neurons in the input and output layers (Figure~\ref{ann_demo}). It must be highlighted that unlike RC-ESN (and RNN-LSTM), this ANN is stateless, i.e., there is no hidden variable such as $r(t)$ that tracks temporal evolution. Furthermore, unlike the RC-ESN, for which the input and output are $X(t)$ and $X(t+\Delta t)$, following Duben and Bauer \cite{dueben2018challenges}, the ANN's inputs/outputs are chosen to be pairs of $X(t)$ and $\Delta X(t)=X(t+\Delta t)-X(t)$ (Figure~\ref{ann_demo}). During prediction, using $X(t)$ that is known from initial condition or previously calculated, $\Delta X(t)$ is predicted. $X(t+\Delta t)$ is then computed via  Adams-Bashforth integration scheme  
\begin{eqnarray}
X(t+\Delta t)=X(t) + \frac{1}{2} \big[ 3\Delta X(t)-\Delta X(t-\Delta t)  \big]. \label{eq:AB}
\end{eqnarray}
{More details on the ANN architecture has been reported in Appendix~\ref{sec:ANN_append}.}




\subsubsection{Training and Prediction}
Training is performed on $N$ pairs of sequential $(X(t),\Delta X(t))$ from the training set. During training, the weights of the network are computed using backpropagation optimized by the stochastic gradient descent algorithm. During prediction, as mentioned above, $\Delta X(t)$ is predicted from $X(t)$ that is known from initial condition or has been previously predicted, and $X(t+\Delta t)$ is then calculated using Eq.~(\ref{eq:AB}).

Our ANN architecture and training/prediction procedure are similar to the ones used in Duben and Bauer~\cite{dueben2018challenges} (we have optimized, by trial and error, the hyperparameters for this particular network). There is no overfitting in the training phase because the final training and testing accuracies are the same. Our code is developed in Keras and is made publicly available (see Code and data availability).

\subsection{Recurrent neural network with long short-term memory (RNN-LSTM)} \label{LSTM}
\subsubsection{Architecture}
The RNN-LSTM \cite{hochreiter1997long} is a deep learning algorithm most suited for prediction of sequential data such as time series, and has received a lot of attention in recent years \cite{goodfellow2016deep}. Variants of RNN-LSTMs are the best performing models for time series modeling in areas such as stock pricing \cite{chen2015lstm}, supply chain \cite{carbonneau2008application}, natural language processing \cite{cho2014learning}, and speech recognition \cite{graves2013speech}. A major improvement over regular RNNs, which have issues with the vanishing and the exploding gradients \cite{pascanu2013difficulty}, LSTMs have become the state-of-the-art approach for training RNNs in the deep learning community. Unlike regular RNNs, the RNN-LSTMs have gates that control the information flow into the neural network from previous time steps of the time series. The RNN-LSTM used here, like the RC-ESN but unlike the ANN, is stateful (i.e., actively maintains state). Our RNN-LSTM has $50$ hidden layers in each RNN-LSTM cell. More details on our RNN-LSTM are presented in \ref{sec:lstm}. RNN-LSTMs have many tunable parameters and are trained with the expensive BPTT algorithm \cite{goodfellow2016deep}. 

\subsubsection{Training and Prediction}
The input to the RNN-LSTM is a time-delay-embedded matrix of $X(t)$ with embedding dimension $q$ (also known as lookback)  \cite{kim1999nonlinear}. An extensive hyperparameter optimization (by trial and error) is performed to find the optimal value of $q$ for which the network has the largest prediction horizon (exploroing $q=1-22$, we found $q=3$ to yield the best performance). The RNN-LSTM predicts $X(t+\Delta t)$ from the previous $q$ time steps of $X(t)$. This is in contrast with RC-ESN, which only uses $X(t)$ and reservoir state $r(t)$ to predict $X(t+\Delta t)$, and ANN, which only uses $X(t)$ and no state to predict $X(t+\Delta t)$ (via predicting $\Delta X(t)$). The weights of the LSTM layers are determined during the training process (see \ref{sec:lstm}). During testing, $X(t+\Delta t)$ is predicted using the past $q$ observables $\left[X(t-(q-1)\Delta t) \dots X(t-\Delta t), X(t)\right]$ that are either known from initial condition or have been previously predicted. We have found the best results with a stateless LSTM (which means that the state of the LSTM gets refreshed during the beginning of each batch during training, see \ref{sec:lstm}) that outperforms a stateful LSTM (where the states gets carried over to the next batch during training).

The architecture of our RNN-LSTM is similar to the one used in Vlachas \textit{et al}.~ \cite{vlachas2018data}. There is no overfitting in the training phase because the final training and testing accuracies are the same. Our code is developed in Keras and is made publicly available (see Code and data availability). 

\section{Results}
\label{sec_results}
\subsection{Short-term prediction: Comparison of the RC-ESN, ANN, and RNN-LSTM performances}
\begin{figure}[t]
\centering
\includegraphics[width=\textwidth]{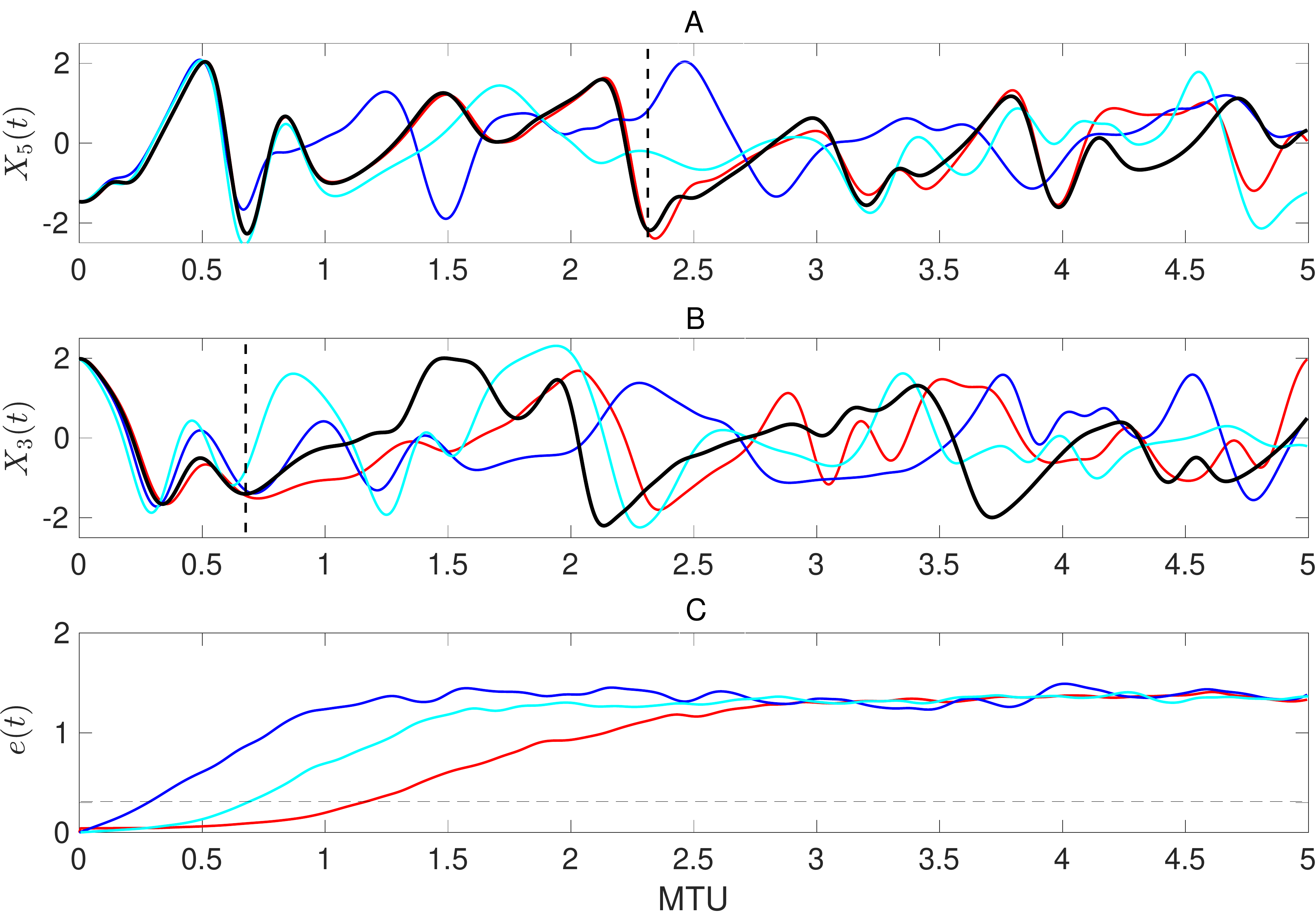}
\caption{Comparison of the short-term forecast skills among the three deep learning methods. The lines show truth (black), RC-ESN (red), ANN (blue), and RNN-LSTM (cyan). (A) and (B) show examples (differing in initial condition) for which RC-ESN yields the longest and shortest prediction horizons, respectively. Vertical dashed lines approximately show the prediction horizon of RC-ESN. (C) Relative $L_2$ error averaged over $100$ randomly chosen initial conditions $e(t)$ (see Eq.~(\ref{error})). The dashed horizontal line marks $e=0.3$. Time is in MTU, where $1$ MTU $=$ $200 \Delta t \approx 4.5/\lambda_{max}$. Training size is $N=500000$.}
\label{esn_time}
\end{figure}

The short-term prediction skills of the three deep learning methods for the same training/testing sets are compared in Figure~\ref{esn_time}. Given the chaotic nature of the system, the performance of the methods depends on the initial condition from which the prediction is conducted. To give the readers a comprehensive view of the performance of these methods, Figures~\ref{esn_time}(A) and (B) show examples of the predicted trajectories (for one element of $X(t)$) versus the true trajectory for two specific initial conditions (from the $100$ initial conditions we used): the one for which RC-ESN shows the best performance (Figure~\ref{esn_time}(A)), and the one for which RC-ESN shows the worst performance (Figure~\ref{esn_time}(B)). 

As seen in Figure~\ref{esn_time}(A), RC-ESN accurately predicts the time series for over $2.3$~MTU, which is equivalent to $460\Delta t$ and over $10.35$ Lyapunov timescales. Closer examination shows that the RC-RSN prediction follows the true trajectory well even up to $\approx 4$~MTU. The RNN-LSTM has the next best prediction performance (up to around $0.9$~MTU or $180 \Delta t$). The prediction from ANN is for around $0.6$~MTU or $120 \Delta t$. For the example in Figure~\ref{esn_time}(B), all methods have shorter prediction horizons, but RC-ESN still has the best performance (accurate prediction up to $\approx 0.7$~MTU), followed by ANN and RNN-LSTM with similar prediction accuracies ($\approx 0.3$~MTU). 

\begin{figure}[t]
\centering
\includegraphics[width=\textwidth]{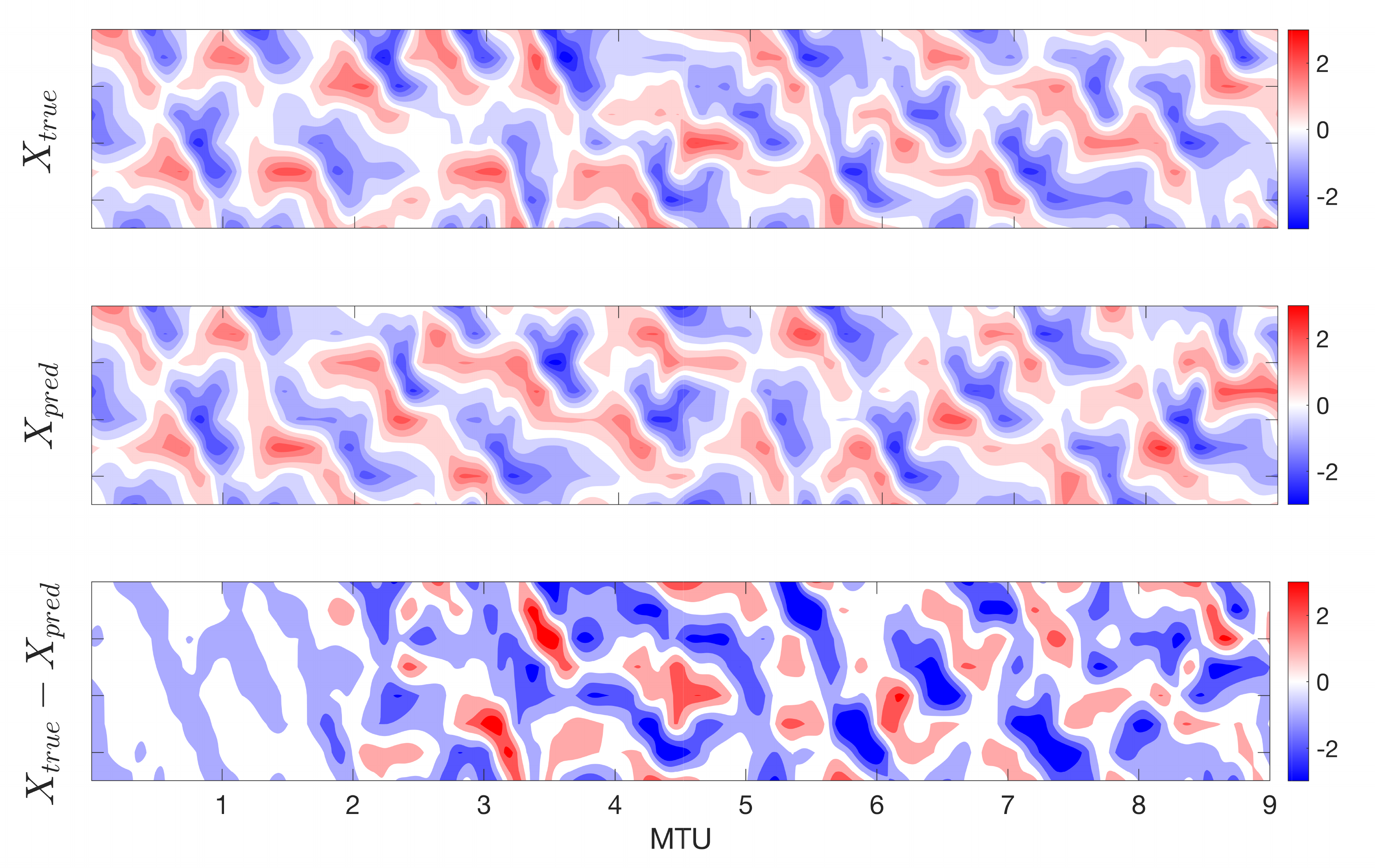}
\caption{Performance of RC-ESN for short-term spatio-temporal prediction. We remind the readers that only the slow/large-scale variable $X$ has been used during training/testing, and the fast/small-scale variables $Y$ and $Z$ have not been used at any point during training or testing. RC-ESN has substantial forecast skills, providing accurate predictions up to $\approx 2$~MTU, which is around $400\Delta t$ or $9$ Lyapunov timescales. $N=500000$, algorithm $T_2$, $D=5000$, and $\rho=0.1$ are used. }
\label{esn_contour}
\end{figure}

Searching through all $100$ initial conditions, the best prediction with RNN-LSTM is up to $\approx 1.7$~MTU (equivalent to $340\Delta t$), and the best prediction with ANN is up to $\approx 1.2$~MTU (equivalent to $240\Delta t$), 

To compare the results over all $100$ randomly chosen initial conditions, we have defined an averaged relative $L_2$ error between the true and predicted trajectories
\begin{eqnarray}
\label{error}
e(t)=\left[\frac{||X_{true}(t)-X_{pred}(t)||}{\left\langle||X_{true}(t)||\right\rangle}\right].
\end{eqnarray} 
Here $\left[ \cdot \right]$ and $\langle \cdot \rangle$ indicate, respectively, averaging over $100$ initial conditions and over $2000 \Delta t$. To be clear, $X_{true}(t)$ refers to the data at time $t$ obtained from the numerical solution while $X_{pred}(t)$ refers to the predicted value at $t$ using one of the deep learning methods. Figure~\ref{esn_time}(C) compares $e(t)$ for the three methods. It is clear that RC-ESN significantly outperforms ANN and RNN-LSTM for short-term prediction in this multi-scale chaotic testbed. Figure~\ref{esn_contour} shows an example of the spatio-temporal evolution of $X_{pred}$ (from RC-ESN), $X_{true}$, and their difference, which further demonstrates the capabilities of RC-ESN for short-term spatio-temporal prediction. 

\subsection{Short-term prediction: Scaling of RC-ESN and ANN performance with training size $N$}
\label{define_E}
How the performance of deep learning techniques scales with the size of the training set is of significant practical importance as the amount of available data is often limited in many problems. Given that currently there is no theoretical understanding of such scaling for these deep learning techniques, we have empirically examined how the quality of short-term predictions scales with $N$. We have conducted the scaling analysis for $N=10^4$ to $N=2 \times 10^6$ for the three methods. Two metrics for the quality of prediction are used: the prediction horizon, defined as when the averaged $L_2$ error $e(t)$ reaches $0.3$, and the prediction error $E$, defined as the average of $e(t)$ between $0-0.5$~MTU:
\begin{eqnarray}
\label{eq:E}
E=\frac{1}{100\Delta t}\sum_{i=0}^{i=100}e(i\Delta t).
\end{eqnarray}
 
\begin{figure}[t]
\centering
\includegraphics[width=\textwidth]{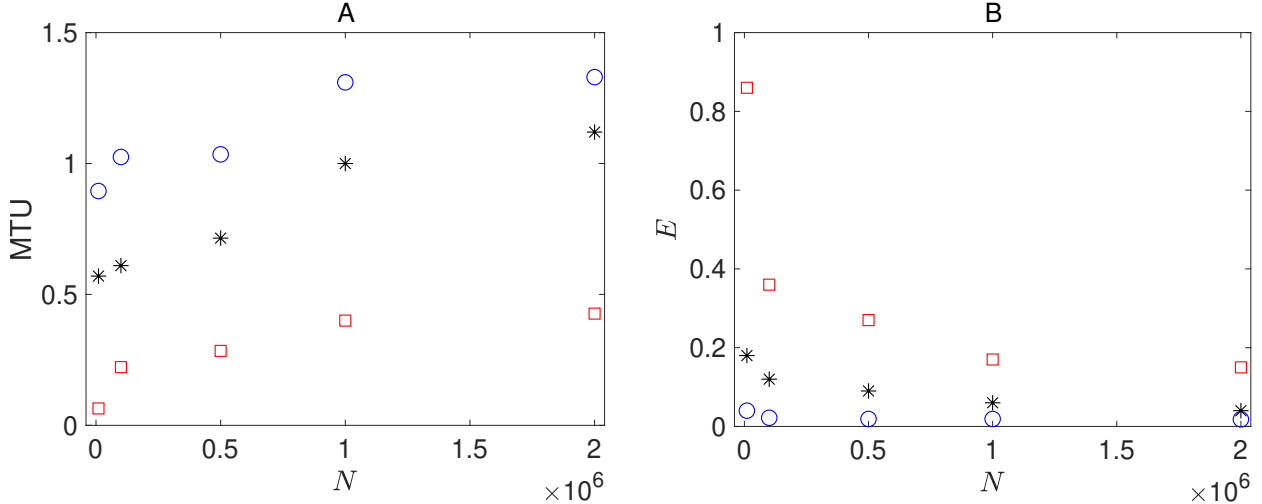}
\caption{Comparison of the short-term prediction quality for RC-ESN (blue circles), RNN-LSTM (black stars), and ANN (red squared) as the size of the training set $N$ is changed from $N=10^4$ to $N=2 \times 10^6$. (A) Prediction horizon (when $e(t)$ reaches $0.3$). (B) Average error $E$ (see Eq.~(\ref{eq:E})). The hyperparameters in each method are optimized for each $N$.}
\label{esn_scaling}
\end{figure}

Figure~\ref{esn_scaling}(A) shows that for all methods, the prediction horizon increases monotonically but nonlinearly as we increase $N$. The prediction horizons of RC-ESN and ANN appear to saturate after $N=10^6$, although the RC-ESN has a more complex step-like scaling curve that needs further examination in future studies. The prediction horizon of RC-ESN exceeds that of ANN by factors ranging from $3$ (for high $N$) to $9$ (for low $N$). In the case of RNN-LSTM, the factor ranges from $1.2$ (for high $N$) to $2$ (for low $N$). Figure~\ref{esn_scaling}(B) shows that for all methods, the average error $E$ decreases as $N$ increases (as expected), most notably for ANN when $N$ is small. 

Overall, compared to both RNN-LSTM and ANN, the prediction horizon and accuracy of RC-ESN have a weaker dependence on the size of the training set, which is a significant advantage for RC-ESN when the dataset available for training is short, which is common in many practical problems.

\subsection{Short-term prediction: Scaling of RC-ESN performance with reservoir size $D$}
Given the superior performance of RC-ESN for short-term prediction, here we focus on one concern with this method: the need for large reservoirs, which can be computationally demanding. This issue has been suggested as a potential disadvantage of ESNs versus LSTMs for training RNNs \cite{Jaeger2007}. Aside from the observations here that RC-ESN significantly outperforms RNN-LSTM for short-term predictions, the problem of reservoir size can be tackled in at least two ways. First, Pathak \textit{et al}.~ \cite{pathak2018model} have proposed, and shown the effectiveness of, using a set of parallelized reservoirs, which allowed for easily dealing with high-dimensional chaotic toy models. 

\begin{figure}[t]
\centering
 \includegraphics[width=0.5\textwidth]{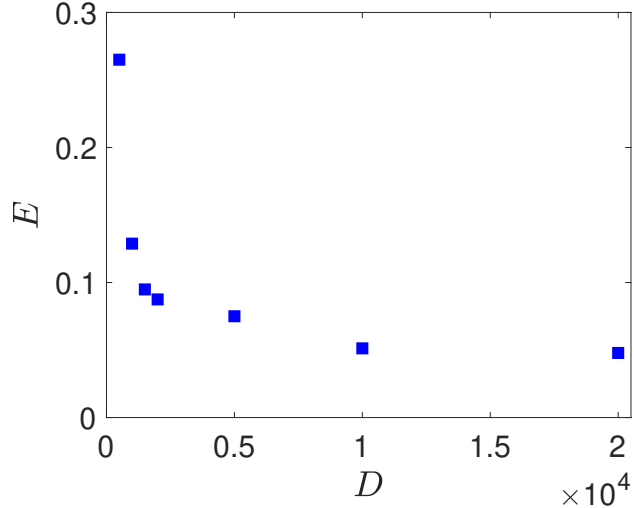}
\caption{For RC-ESN: scaling of average prediction error between $0-0.5$~MTU ($E$,  Eq.~(\ref{eq:E})) with reservoir size $D$. $N=500000$ is used.}
\label{res_scaling}
\end{figure}

Second, Figure~\ref{res_scaling} shows that $E$ rapidly declines by a factor of around $3$ as $D$ is increased from $500$ to $5000$, decreases slightly as $D$ is further doubled to $10000$, and then barely changes as $D$ is doubled again to $20000$. Training the RC-ESN with $D=20000$ versus $D=5000$ comes with a higher computational cost ($16$ GB memory and $\approx 6$ CPU hours for $D=5000$ and $64$ GB memory and $\approx 18$ CPU hours for $D=20000$), while little improvement in accuracy is gained. Thus, concepts from inexact computing can be used to choose $D$ such that precision is traded for large savings in computational resources, which can be then reinvested into more simulations, higher resolutions for critical processes etc. \cite{palem2014inexactness,palmer2014climate,duben2014benchmark,leyffer2016doing,thornes2017use}. 

\subsection{Long-term statistics: Comparison of RC-ESN, ANN, and RNN-LSTM performance}
All the data-driven predictions discussed earlier eventually diverge from the true trajectory (as would even predictions using the numerical solver). Still, it is interesting to examine whether the freely predicted spatio-temporal data have the same long-term statistical properties as the actual dynamical system (i.e., Eqs.~(\ref{eq:L1})-(\ref{eq:L3})). Reproducing the actual dynamical system's long-term statistics (sometimes refer to as the system's \emph{climate}) using a data-driven method can be significantly useful. In some problems, a surrogate model does not need to predict the evolution of a specific trajectory, but only the long-term statistics of a system. Furthermore, synthetic long datasets produced using an inexpensive data-driven method (trained on a shorter real dataset) can be used to examine the system's probability density functions (PDFs), including its tails, which are important for studying the statistics of rare/extreme events.

By examining return maps, Pathak \textit{et al}.~\cite{pathak2017using} have already shown that RC-ESNs can reproduce the long-term statistics of the Lorenz 63  and {Kuramoto-Sivashinsky} equations (Jaeger\textit{et al}.~\cite{jaeger2004harnessing} and Pathak \textit{et al}.~\cite{pathak2017using}  have also shown that RC-ESNs can be used to accurately estimate a chaotic system's Lyapunov spectrum). Here, we focus on comparing the performance of RC-ESN, ANN, and RNN-LSTM in reproducing the system's long-term statistics by
\begin{itemize}
\item Examining the estimated PDFs and in particular their tails,
\item Investigating whether the quality of the estimated PDFs degrades with time, which can negate the usefulness of long synthetic datasets.
\end{itemize}

\begin{figure}[t]
\centering
 \includegraphics[width=\textwidth]{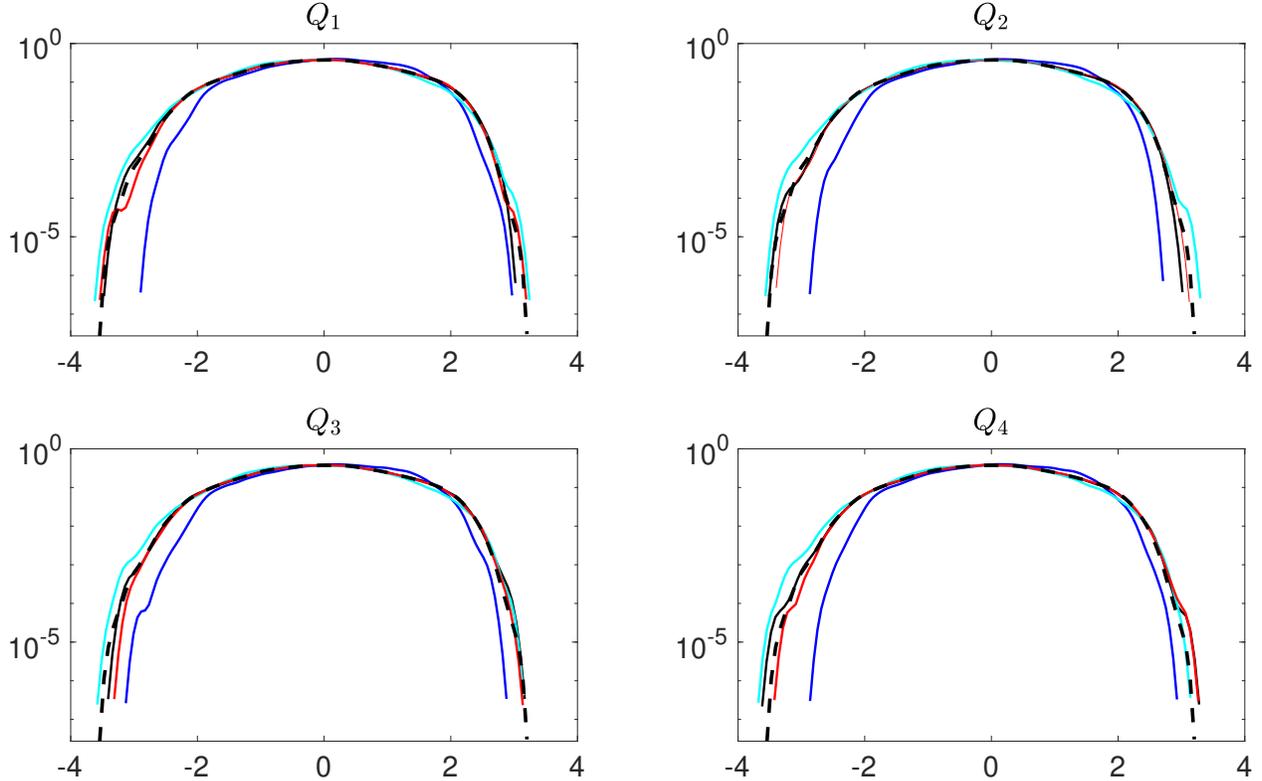}
\caption{Estimated probability density functions (PDFs) of long datasets freely predicted using RC-ESN (red), ANN (blue), and RNN-LSTM (cyan) compared to the true PDF (dashed black). RC-ESN, ANN, and RNN-LSTM are first trained with $N=500000$, and then predict, from an initial condition, for $4 \times 10^6 \Delta t$. The panels $Q_1-Q_4$ correspond to the equally divided quartiles of $10^6$ predicted samples. PDFs are approximated using kernel density estimation \cite{epanechnikov1969non}. The green lines show the estimated PDFs from different sets of $10^6$ samples that are obtained from numerical solution of Eqs.~(\ref{eq:L1})-(\ref{eq:L3}). Small differences between the green lines show the uncertainties in estimating the tails from $10^6$ samples. The dashed black lines show the true PDF estimated from  $10^7$ samples from numerical solution. Note that the presented PDFs are for standardized data.}
\label{stat}
\end{figure}

Figure~\ref{stat} compares the estimated PDFs obtained using the three deep learning methods. The data predicted using RC-ESN and RNN-LSTM are found to have PDFs closely matching the true PDF, even at the tails. Deviations at the tails of the PDF predicted by these methods from the true PDF are comparable to the deviations of the PDFs obtained from true data using the same number of samples. The ANN-predicted data have reasonable PDFs between $\pm 2$ standard deviations, but the tails have substantial deviations from those of the true PDFs. All predicted PDFs are robust and do not differ much (except near the end of the tails) among the quartiles.

The results show that RC-ESN and RNN-LSTM can accurately reproduce the system's long-term statistics, and can robustly produce long synthetic datasets with PDFs that are close to the PDF of the true data even near the tails. The ability of ANN to accomplish these tasks is limited. 


\section{Discussion}\label{sec:discussion}
By examining the true and predicted trajectories (Figures~\ref{esn_time}-\ref{esn_contour}) and the prediction errors and horizons (Figure~\ref{esn_scaling}), we have shown that RC-ESN substantially outperforms ANN and RNN-LSTM in predicting the short-term evolution of a multi-scale Lorenz 96 chaotic system (Eqs.~(\ref{eq:L1})-(\ref{eq:L3})). Additionally, RC-ESN and RNN-LSTM both work well in reproducing the long-term statistics of this system (Figure~\ref{stat}). We emphasize that following the problem formulation of Duben and Bauer \textit{et al}.~\cite{dueben2018challenges}, and unlike most other studies, only part of the multi-scale state-vector (the slow/large-scale variable $X$) has been available for training the data-driven model and has been of interest for testing. This problem design is more relevant to many practical problems but is more challenging as it requires the data-driven model to not only predict the evolution of $X$ based on its past observations, but also to account for the effects of the intermediate and fast/small-scale variables $Y$ and $Z$.

We have also found that the prediction horizon, and in particular the prediction accuracy, of RC-ESN to have a weak dependence on the size of the training set  (Figure~\ref{esn_scaling}). This is an important property, as in many practical problems the data available for training are limited. Furthermore, the prediction error of RC-ESN is shown to have an asymptotic behavior for large reservoir sizes, which suggests that reasonable accuracy can be achieved with moderate reservoir sizes. Note that the skillful predictions with RC-ESN in Figures~\ref{esn_time}-\ref{esn_contour} have been obtained with a moderate-sized training set ($N=500000$) and reservoir ($D=5000$). Figures~\ref{esn_scaling} and \ref{res_scaling} suggest that slightly better results could have been obtained using larger $N$ and $D$, although such improvements come with a higher computational cost during training.

The order we have found for the performance of the three deep learning methods (RC-ESN $>$ RNN-LSTM $>$ ANN) is different from the order of complexity of these methods in terms of the number of trainable parameters (RC-ESN $<$ ANN $<$ RNN-LSTM) but aligned with the order in terms of the learnable function space (ANN $<$ RC-ESN and RNN-LSTM, as the latter methods are both RNNs and thus Turning complete, while ANN is a state-less feed-forward network and thus not Turing complete \cite{siegelmann1992computational}). {Whether the superior predictive performance of RC-ESN (especially over RNN-LSTM) is due to its explicit state representation updated at every time step both during training and testing, or lower likelihood of overfitting due to the order-of-magnitude smaller number of trainable parameters, remains to be seen.} Also whether we have not fully harnessed the power of RNN-LSTMs (see below) is unclear at this point, particularly because a complete theoretical understanding of how/why these methods work (or do not work) is currently lacking. That said, there have been some progress in understanding the RC-ESNs, in particular by modeling the network itself as a dynamical system \cite{yildiz2012re,gauthier2018reservoir}. Such efforts, for example those aimed at understanding the echo states that are learned in the RC-ESN's reservoir, might benefit from recent advances in dynamical systems theory \cite{mezic2005spectral,tu2014dynamic,williams2015data,arbabi2017ergodic,giannakis2017spatiotemporal,khodkar2018data}.    

\emph{An important next step in our work is to determine how generalizable our findings are}. This investigation is important for the following two reasons. First, here we have only studied one system, a specially designed version of the Lorenz 96 system. The performance of these methods should be examined in a hierarchy of chaotic dynamical systems and high-dimensional turbulent flows. That said, our findings are overall consistent with the recently reported performance of these methods applied to chaotic toy models. \cite{pathak2018model,pathak2017using,lu2017reservoir} demonstrated that RC-ESN can predict, for several Lyapunov timescales, the spatio-temporal evolution, and can reproduce the climate of the Lorenz 63 and Kuramoto-Sivashinsky chaotic systems. Here, we have shown that RC-ESN performs similarly well even when only the slow/large-scale component of the multi-scale state-vector is known. Our results with ANN are consistent with those of Dueben and Bauer~\cite{dueben2018challenges}, who showed examples of trajectories predicted accurately up to $1$~MTU with a large training set $N=2 \times 10^6$ using ANN for the same Lorenz 96 system (see their Fig.~1). 
While RNN-LSTM is considered state-of-the-art for sequential data modeling, and has worked well for a number of applications involving time series, to the best of our knowledge, simple RNN-LSTMs, such as the one used here, have not been very successful when applied to chaotic dynamical systems. Vlachas \textit{et al}.~\cite{vlachas2018data} found some prediction skills using RNN-LSTM applied to the original Lorenz 96 system (which does not have the multi-scale coupling); see their Fig.~5.  

Second, we have only used a simple RNN-LSTM; there are other variants of this architecture as well as more advanced deep learning RNNs that might potentially yield better results. For our simple RNN-LSTM, we have extensively explored optimization of hyperparameters and tried variant formulations of the problem: predict $\Delta X(t)$ with or without updating the state of the LSTM. We have found that such variants do not lead to better results, consistent with the findings of Vlachas \textit{et al}.~\cite{vlachas2018data}. Our preliminary explorations with more advanced variants of RNN-LSTM (seq2seq and encoder-decoder LSTM \cite{sutskever2014sequence}) have not resulted in any improvement either. However, just as the skillful predictions of RC-ESN and ANN hinge on one key step (nonlinear transformation for the former and predicting $\Delta X$ for the latter), it is possible that changing/adding one step leads to major improvements in RNN-LSTM {(it is worth mentioning that similar nonlinear transformation of the state in LSTMs is not straightforward due to their more complex architecture; see Appendix~\ref{sec:lstm})}. We have shared our codes publicly to help others explore such changes to our RNN-LSTM. Furthermore, there are other more sophisticated RNNs that we have not explored. For example \cite{yu2017long} have introduced a tensor-train RNN that outperformed a simple RNN-LSTM in predicting the temporal evolution of the Lorenz 63 chaotic system. Mohan \textit{et al.}.~\cite{mohan2019compressed} showed that a compressed convolutional LSTM performs well in capturing the long-term statistics of a turbulent flow. The performance of more advanced deep learning RNNs should be examined and compared, side by side, with the performance of RC-ESNs and ANNs in future studies. The multi-scale Lorenz 96 system provides a suitable starting point for such comparisons.

\emph{The results of our work show the promise of deep learning methods such as RC-ESNs (and to some extent RNN-LSTM) for data-driven modeling of chaotic dynamical systems, which has broad applications in geosciences, e.g., in weather/climate modeling}. Practical and fundamental issues such as interpretability, scalability to higher dimensional systems \cite{pathak2018model}, presence of measurement noise in the training data and initial conditions \cite{rudy2018deep}, non-stationarity of the time series, and dealing with data that have two or three spatial dimensions (e.g., through integration with convolutional neural networks, CNN-LSTM \cite{xingjian2015convolutional} and CNN-ESN \cite{ma2017walking} should be studied in future work. 

Here we have focused on a fully data-driven approach, as opposed to the conventional approach of direct numerical solutions (Figure~\ref{data_drive_demo}). In practice, for example for large-scale, multi-physics, multi-scale dynamical systems such as weather and climate models, it is likely that a hybrid framework yields the best performance: depending on the application and the spatio-temporal scales of the physical processes involved \cite{thornes2017use,chantry2019scale}, some of the equations could be solved numerically with double precision, some could be solved numerically with lower precisions, and some could be approximated with a surrogate model learned via a data-driven approach, such as the ones studied in this paper.  


\appendix

\section{More details on RC-ESN}\label{sec:appnT}
Here we show a comparison of RC-ESN forecast skills with nonlinear transformation algorithms $T_1$ \cite{pathak2018model}, $T_2$ and $T_3$, and without any transformation between $r(t)$ and $\tilde{r}(t)$. The three algorithms are (for $i=1,2,3 \dots N$ and $j=1,2,3 \dots D$) \\

\noindent {\bf Algorithm $T_1$} \\
$\tilde{r}_{i,j} =r_{i,j} \times r_{i,j}$ \space \space  \space \space  \space   if $j$ is odd \\
$\tilde{r}_{i,j} =r_{i,j}$ \space \space  \space \space \space  \space  \space  \space  \space  \space  \space  \space  \space   \space \space  \space if $j$ is even \\

\noindent {\bf Algorithm $T_2$} \\
$\tilde{r}_{i,j} =r_{i,j-1} \times r_{i,j-2}$ \space \space  \space \space  \space  if $j >1$ is odd \\
$\tilde{r}_{i,j} =r_{i,j}$ \space \space  \space \space \space  \space  \space  \space  \space  \space \space  \space \space \space  \space  \space  \space  \space  \space  \space  \space  \space   \space \space  \space  if $j$ is $1$ or even\\

\noindent {\bf Algorithm $T_3$} \\
$\tilde{r}_{i,j} =r_{i,j-1} \times r_{i,j+1}$ \space \space  \space \space  \space if $j>1$ is odd \\
$\tilde{r}_{i,j} =r_{i,j}$ \space \space  \space \space \space  \space  \space  \space  \space \space  \space \space \space  \space  \space  \space  \space  \space  \space  \space  \space \space \space  \space \space  if $j$ is $1$ or even \\

\begin{figure}[t]
\centering
\includegraphics[width=\textwidth]{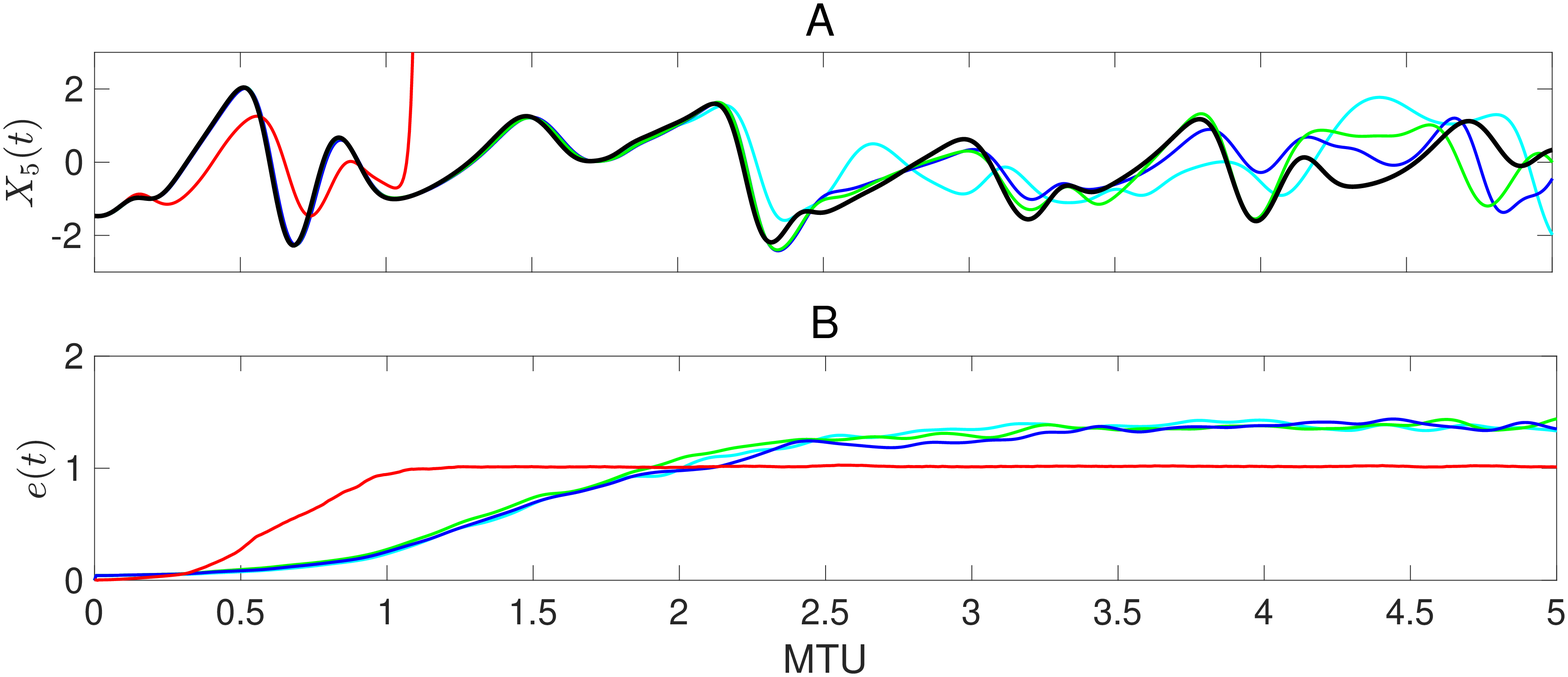}
\caption{Comparison of RC-ESN forecast skills with nonlinear transformation algorithms $T_1$, $T_2$, and $T_3$ (cyan, green, and blue lines, respectively) and without any transformation (red line). Black line shows the truth. (A) Prediction from one initial condition, (B) Relative $L_2$-norm error averaged over $100$ initial condition $e(t)$ (Eq.~(\ref{error})). Time is in MTU where $1$ MTU $=$ $200 \Delta t \approx 4.5/\lambda_{max}$.}
\label{esn_basis}
\end{figure}

Figure~\ref{esn_basis}(A) shows an example of short-term predictions from an initial condition using $T_1$, $T_2$, $T_3$, and no transformation, everything else kept the same. It is clear that the nonlinear transformation is essential for skillful predictions, as the prediction obtained without transformation diverges from the truth before $0.25$~MTU. The three nonlinear transformation algorithms yield similar results, with accurate predictions for more than $2$~MTU. Figure~\ref{esn_basis}(B), which shows the relative prediction error averaged over $100$ initial conditions, further confirms this point.

Why the nonlinear transformation is needed, and the best/optimal choice for the transformation (if it exists) should be studied in future work. We highlight that the nonlinear transformation resembles basis function expansion, which is commonly used to capture nonlinearity in linear regression models \cite{bishop2006pattern}.

\section{More details on ANN}\label{sec:ANN_append}
{The ANN used in this study (and the one that yields the best performance) is similar to that of Duben and Bauer \textit{et al}.~\cite{dueben2018challenges}. The ANN has four hidden layers each of which has $100$ neurons with a $\tanh$ activation function. The input to the ANN has $8$ neurons which takes $X(t)$ as the input and outputs $\Delta X(t)$ which is also a $8$-neuron layer. The obtained $\Delta X(t)$ output is then used in an Adam-Bashforth integration scheme to obtain $X(t+\Delta t)$. We found that a simpler Euler scheme yields similar results. However, we found that training/testing on pairs of $X(t)$ and $X(t+\Delta t$) (that was used for RC-ESN) leads to no prediction skill with ANN, and that following the procedure used in Dueben and Bauer~\cite{dueben2018challenges} is essential for skillful predictions. We speculate that this might be due to the stateless nature of the ANN. By relating $X(t)$ to the change in $X(t)$ at the next time step, rather than the raw value $X(t+\Delta t)$, this ANN training architecture implicitly contains a reference to the previous time step. While the ANN itself is stateless, this particular training approach essentially encodes a first-order temporal dependence between successive states.}

{Note that our approach here is the same as the global ANN of Duben and Bauer \textit{et al}.~ \cite{dueben2018challenges}. We also tried their local ANN approach, but consistent with their findings for the Lorenz system, found better performance with the global approach (results not reported for brevity).} 

{The ANN is trained with a stochastic gradient descent algorithm with a learning rate of $0.001$ with a batch size of $100$ and mean absolute error as loss function (mean squared error also gives similar performance). 
}

\section{More details on RNN-LSTM}\label{sec:lstm}
The governing equations for RNN-LSTM are:
\begin{eqnarray}
g^f(t)=\sigma_f\left( \mathbf{W_f}\left[h(t-1), I(t)\right] +b_f \right),\\
g^i(t)=\sigma_f\left( \mathbf{W_i}\left[h(t-1), I(t)\right] +b_i \right),\\
\tilde{C}(t)=\tanh\left(\mathbf{W_c} \left[h(t-1), I(t) \right] +b_h \right),\\
C(t)=g^f(t) C(t-1)+g^i(t) \tilde{C}(t),\\
g^o(t)=\sigma_h\left( \mathbf{W_h}\left[h(t-1), I(t)\right] +b_h \right),\\
h(t)=g^o(t) \tanh\left( C(t)\right),\\
\Phi(t)=\mathbf{W_{oh}}h(t), \\
X(t+\Delta t) \approx \Phi(t).
\end{eqnarray}
$\sigma_f$ is the softmax activation function; $g^f(t)$, $g^i(t)$, and $g^o(t)$ $\in \Re^{d_h \times \left( d_h+d_i\right)}$ are the forget gate, input gate, and output gate respectively. $d_h$ is the dimension of the hidden layers (chosen as $50$ in our study) and $d_i$ is the dimension of the input ($8 \times q$). $b_f$, $b_i$, and $b_h$ are the biases in the forget gate, input gate, and the hidden layers. $I(t) \in \Re^{d_i}$ is the input, which is a column of a time-delay-embedded matrix of $X(t)$.
This matrix has the dimension of $(8 \times q) \times N$. $h(t) \in \Re^{d_h}$ is the hidden state and $C(t) \in \Re^{d_h}$ is the cell state (the states track the temporal evolution). The weights $\mathbf{W_o}$, $\mathbf{W_i}$, $\mathbf{W_f}$, $\mathbf{W_c}$, and $\mathbf{W_{oh}}$ are learned through the BPTT algorithm \cite{goodfellow2016deep} using an ADAM optimizer \cite{kingma2014adam}. $\Phi(t)$ is the output from the RNN-LSTM. 

The LSTM used in this study is a stateless LSTM, wherein the two hidden states ($C$ and $h$) are refreshed at the beginning of each batch during training (this is not the same as the stateless ANN, where there is no hidden state at all). Here, the training sequences in each batch are shuffled randomly, leading to an unbiased gradient estimator in the stochastic gradient descent algorithm \cite{meng2019convergence}.\\

{\bf Code and data availability:}All data and codes can be accessed at \url{https://github.com/ashesh6810/RCESN_spatio_temporal.}\\

{\bf Acknowledgment:} This work was supported by NASA grant 80NSSC17K0266 and an Early-Career Research Fellowship from the Gulf Research Program of the National Academies of Science, Engineering, and Medicine (to P.H.). A.C. thanks the Rice University Ken Kennedy Institute for a BP HPC Graduate Fellowship. Computational resources on Bridge Large, Bridge GPU, and Stampede2 clusters were provided by the NSF XSEDE allocation ATM170020. We are grateful to Krishna Palem for many stimulating discussions and helpful comments. We thank Michelle Girvan, Reza Malek-Madani, Rohan Mukherjee, Guido Novati, and Pantelis Vlachas for insightful discussions, and Jaideep Pathak for sharing his Matlab ESN code for the Kuramoto-Sivashinsky system. We are grateful to Mingchao Jiang and Adam Subel for help with running some of the simulations. 







 \bibliographystyle{unsrt}
 \bibliography{template}

\begin{thebibliography}{10}

\bibitem{collins2006formulation}
W.D. Collins, P.J. Rasch, B.A. Boville, J.J. Hack, J.R. McCaa, D.L. Williamson,
  B.P. Briegleb, C.M. Bitz, S.J. Lin, and M.~Zhang.
\newblock The formulation and atmospheric simulation of the community
  atmosphere model version 3 ({CAM}3).
\newblock {\em Journal of Climate}, 19(11):2144--2161, 2006.

\bibitem{collins2011development}
W.J. Collins et~al.
\newblock Development and evaluation of an {Earth-System model}--{HadGEM2}.
\newblock {\em Geoscientific Model Development}, 4(4):1051--1075, 2011.

\bibitem{flato2011earth}
G.M. Flato.
\newblock Earth system models: an overview.
\newblock {\em Wiley Interdisciplinary Reviews: Climate Change}, 2(6):783--800,
  2011.

\bibitem{bauer2015quiet}
P.~Bauer, A.~Thorpe, and G.~Brunet.
\newblock The quiet revolution of numerical weather prediction.
\newblock {\em Nature}, 525(7567):47, 2015.

\bibitem{jeevanjee2017perspective}
N.~Jeevanjee, P.~Hassanzadeh, S.~Hill, and A.~Sheshadri.
\newblock A perspective on climate model hierarchies.
\newblock {\em Journal of Advances in Modeling Earth Systems}, 9(4):1760--1771,
  2017.

\bibitem{stevens2013climate}
B.~Stevens and S.~Bony.
\newblock What are climate models missing?
\newblock {\em Science}, 340(6136):1053--1054, 2013.

\bibitem{hourdin2017art}
F.~Hourdin, Thorsten M., Andrew G., J.C. Golaz, V.~Balaji, Q.~Duan, D.~Folini,
  D.~Ji, D.~Klocke, Y.~Qian, et~al.
\newblock The art and science of climate model tuning.
\newblock {\em Bulletin of the American Meteorological Society},
  98(3):589--602, 2017.

\bibitem{garcia2017modification}
R.R. Garcia, A.K. Smith, D.E. Kinnison, {\'A}.~la C{\'a}mara, and D.J. Murphy.
\newblock Modification of the gravity wave parameterization in the {Whole
  Atmosphere Community Climate Model}: Motivation and results.
\newblock {\em Journal of the Atmospheric Sciences}, 74(1):275--291, 2017.

\bibitem{schneider2017climate}
T.~Schneider, J.~Teixeira, C.S. Bretherton, F.~Brient, K.G. Pressel,
  C.~Sch{\"a}r, and A.P. Siebesma.
\newblock Climate goals and computing the future of clouds.
\newblock {\em Nature Climate Change}, 7(1):3, 2017.

\bibitem{khairoutdinov2001cloud}
M.F. Khairoutdinov and D.A. Randall.
\newblock A cloud resolving model as a cloud parameterization in the {NCAR
  Community Climate System Model}: Preliminary results.
\newblock {\em Geophysical Research Letters}, 28(18):3617--3620, 2001.

\bibitem{benedict2009structure}
J.J. Benedict and D.A. Randall.
\newblock Structure of the {Madden--Julian} oscillation in the
  superparameterized {CAM}.
\newblock {\em Journal of the Atmospheric Sciences}, 66(11):3277--3296, 2009.

\bibitem{andersen2012moist}
J.A. Andersen and Z.~Kuang.
\newblock Moist static energy budget of {MJO}-like disturbances in the
  atmosphere of a zonally symmetric aquaplanet.
\newblock {\em Journal of Climate}, 25(8):2782--2804, 2012.

\bibitem{kooperman2018rainfall}
G.J. Kooperman, M.S. Pritchard, T.A. O'Brien, and B.W. Timmermans.
\newblock Rainfall from resolved rather than parameterized processes better
  represents the present-day and climate change response of moderate rates in
  the community atmosphere model.
\newblock {\em Journal of Advances in Modeling Earth Systems}, 10(4):971--988,
  2018.

\bibitem{palem2014inexactness}
K.V. Palem.
\newblock Inexactness and a future of computing.
\newblock {\em Philosophical Transactions of the Royal Society A: Mathematical,
  Physical and Engineering Sciences}, 372(2018):20130281, 2014.

\bibitem{palmer2014climate}
T.N. Palmer.
\newblock Climate forecasting: Build high-resolution global climate models.
\newblock {\em Nature News}, 515(7527):338, 2014.

\bibitem{duben2014use}
P.D. D{\"u}ben, J.~Joven, A.~Lingamneni, H.~McNamara, G.~De~Micheli, K.V.
  Palem, and T.N. Palmer.
\newblock On the use of inexact, pruned hardware in atmospheric modelling.
\newblock {\em Philosophical Transactions of the Royal Society A: Mathematical,
  Physical and Engineering Sciences}, 372(2018):20130276, 2014.

\bibitem{duben2014benchmark}
P.D. D{\"u}ben and T.N. Palmer.
\newblock Benchmark tests for numerical weather forecasts on inexact hardware.
\newblock {\em Monthly Weather Review}, 142(10):3809--3829, 2014.

\bibitem{thornes2017use}
T.~Thornes, P.~D{\"u}ben, and T.~Palmer.
\newblock On the use of scale-dependent precision in {E}arth system modelling.
\newblock {\em Quarterly Journal of the Royal Meteorological Society},
  143(703):897--908, 2017.

\bibitem{hatfield2018improving}
S.~Hatfield, A.~Subramanian, T.~Palmer, and P/~D{\"u}ben.
\newblock Improving weather forecast skill through reduced-precision data
  assimilation.
\newblock {\em Monthly Weather Review}, 146(1):49--62, 2018.

\bibitem{duben2015opportunities}
P.~D{\"u}ben, S.~Yenugula, J.~Augustine, K.~Palem, J.~Schlachter, C.~Enz, and
  T.N. Palmer.
\newblock Opportunities for energy efficient computing: A study of inexact
  general purpose processors for high-performance and big-data applications.
\newblock In {\em 2015 Design, Automation \& Test in Europe Conference \&
  Exhibition (DATE)}, pages 764--769. IEEE, 2015.

\bibitem{leyffer2016doing}
S.~Leyffer, S.M. Wild, M.~Fagan, M.~Snir, K.V. Palem, K.~Yoshii, and H.~Finkel.
\newblock Doing {M}oore with less--leapfrogging {M}oore's law with inexactness
  for supercomputing.
\newblock {\em arXiv preprint arXiv:1610.02606}, 2016.

\bibitem{schneider2017earth}
T.~Schneider, S.~Lan, A.~Stuart, and J.~Teixeira.
\newblock Earth system modeling 2.0: A blueprint for models that learn from
  observations and targeted high-resolution simulations.
\newblock {\em Geophysical Research Letters}, 44(24), 2017.

\bibitem{kutz2017deep}
J.N. Kutz.
\newblock Deep learning in fluid dynamics.
\newblock {\em Journal of Fluid Mechanics}, 814:1--4, 2017.

\bibitem{gentine2018could}
P.~Gentine, M.~Pritchard, S.~Rasp, G.~Reinaudi, and G.~Yacalis.
\newblock Could machine learning break the convection parameterization
  deadlock?
\newblock {\em Geophysical Research Letters}, 45(11):5742--5751, 2018.

\bibitem{moosavi2018machine}
Azam Moosavi, Ahmed Attia, and Adrian Sandu.
\newblock A machine learning approach to adaptive covariance localization.
\newblock {\em arXiv preprint arXiv:1801.00548}, 2018.

\bibitem{wu2019enforcing}
J.~Wu et~al.
\newblock Enforcing statistical constraints in generative adversarial networks
  for modeling chaotic dynamical systems.
\newblock {\em arXiv preprint arXiv:1905.06841}, 2019.

\bibitem{toms2019deep}
Benjamin~A Toms, Karthik Kashinath, Da~Yang, et~al.
\newblock Deep learning for scientific inference from geophysical data: The
  {Madden--Julian} oscillation as a test case.
\newblock {\em arXiv preprint arXiv:1902.04621}, 2019.

\bibitem{brunton2019data}
S.L. Brunton and J.N. Kutz.
\newblock {\em Data-driven Science and Engineering: Machine Learning, Dynamical
  Systems, and Control}.
\newblock Cambridge University Press, 2019.

\bibitem{duraisamy2019turbulence}
K.~Duraisamy, G.~Iaccarino, and H.~Xiao.
\newblock Turbulence modeling in the age of data.
\newblock {\em Annual Review of Fluid Mechanics}, 51:357--377, 2019.

\bibitem{reichstein2019deep}
M.~Reichstein et~al.
\newblock Deep learning and process understanding for data-driven {E}arth
  system science.
\newblock {\em Nature}, 566(7743):195, 2019.

\bibitem{lim2019predicting}
Soon~Hoe Lim, Ludovico~Theo Giorgini, Woosok Moon, and JS~Wettlaufer.
\newblock Predicting rare events in multiscale dynamical systems using machine
  learning.
\newblock {\em arXiv preprint arXiv:1908.03771}, 2019.

\bibitem{scher2019generalization}
Sebastian Scher and Gabriele Messori.
\newblock Generalization properties of feed-forward neural networks trained on
  {L}orenz systems.
\newblock {\em Nonlinear Processes in Geophysics}, 26(4):381--399, 2019.

\bibitem{rasp2018deep}
S.~Rasp, M.S. Pritchard, and P.~Gentine.
\newblock Deep learning to represent subgrid processes in climate models.
\newblock {\em Proceedings of the National Academy of Sciences},
  115(39):9684--9689, 2018.

\bibitem{brenowitz2018prognostic}
N.D. Brenowitz and C.S. Bretherton.
\newblock Prognostic validation of a neural network unified physics
  parameterization.
\newblock {\em Geophysical Research Letters}, 45(12):6289--6298, 2018.

\bibitem{gagne2019machine}
II~Gagne, David John, Hannah~M Christensen, Aneesh~C Subramanian, and Adam~H
  Monahan.
\newblock Machine learning for stochastic parameterization: Generative
  adversarial networks in the {L}orenz'96 model.
\newblock {\em arXiv preprint arXiv:1909.04711}, 2019.

\bibitem{o2018using}
P.A. O'Gorman and J.G. Dwyer.
\newblock Using machine learning to parameterize moist convection: Potential
  for modeling of climate, climate change, and extreme events.
\newblock {\em Journal of Advances in Modeling Earth Systems},
  10(10):2548--2563, 2018.

\bibitem{bolton2019applications}
T.~Bolton and L.~Zanna.
\newblock Applications of deep learning to ocean data inference and subgrid
  parameterization.
\newblock {\em Journal of Advances in Modeling Earth Systems}, 11(1):376--399,
  2019.

\bibitem{dueben2018challenges}
P.D. D{\"u}eben and P.~Bauer.
\newblock Challenges and design choices for global weather and climate models
  based on machine learning.
\newblock {\em Geoscientific Model Development}, 11(10):3999--4009, 2018.

\bibitem{watson2019applying}
P.A.G. Watson.
\newblock Applying machine learning to improve simulations of a chaotic
  dynamical system using empirical error correction.
\newblock {\em Journal of Advances in Modeling Earth Systems}, 2019.

\bibitem{salehipour2019deep}
H.~Salehipour and W.R. Peltier.
\newblock Deep learning of mixing by two atoms of stratified turbulence.
\newblock {\em Journal of Fluid Mechanics}, 861, 2019.

\bibitem{ling2016reynolds}
J.~Ling, A.~Kurzawski, and J.~Templeton.
\newblock Reynolds averaged turbulence modelling using deep neural networks
  with embedded invariance.
\newblock {\em Journal of Fluid Mechanics}, 807:155--166, 2016.

\bibitem{mcdermott2017ensemble}
P.L. McDermott and C.K. Wikle.
\newblock An ensemble quadratic echo state network for non-linear
  spatio-temporal forecasting.
\newblock {\em Stat}, 6(1):315--330, 2017.

\bibitem{pathak2018model}
J.~Pathak, B.~Hunt, M.~Girvan, Z.~Lu, and E.~Ott.
\newblock Model-free prediction of large spatiotemporally chaotic systems from
  data: A reservoir computing approach.
\newblock {\em Physical Review Letters}, 120(2):024102, 2018.

\bibitem{rudy2018deep}
S.H. Rudy, J.N. Kutz, and S.L. Brunton.
\newblock Deep learning of dynamics and signal-noise decomposition with
  time-stepping constraints.
\newblock {\em arXiv preprint arXiv:1808.02578}, 2018.

\bibitem{vlachas2018data}
P.R. Vlachas, W.~Byeon, Z.Y. Wan, T.P. Sapsis, and P.~Koumoutsakos.
\newblock Data-driven forecasting of high-dimensional chaotic systems with long
  short-term memory networks.
\newblock {\em Proceedings of the Royal Society A: Mathematical, Physical and
  Engineering Sciences}, 474(2213):20170844, 2018.

\bibitem{mohan2019compressed}
A.~Mohan, D.~Daniel, M.~Chertkov, and D.~Livescu.
\newblock Compressed convolutional {LSTM}: An efficient deep learning framework
  to model high fidelity {3D} turbulence.
\newblock {\em arXiv preprint arXiv:1903.00033}, 2019.

\bibitem{raissi2019physics}
M.~Raissi, P.~Perdikaris, and G.E. Karniadakis.
\newblock Physics-informed neural networks: A deep learning framework for
  solving forward and inverse problems involving nonlinear partial differential
  equations.
\newblock {\em Journal of Computational Physics}, 378:686--707, 2019.

\bibitem{zhu2019physics}
Y.~Zhu, N.~Zabaras, P.S. Koutsourelakis, and P.~Perdikaris.
\newblock Physics-constrained deep learning for high-dimensional surrogate
  modeling and uncertainty quantification without labeled data.
\newblock {\em arXiv preprint arXiv:1901.06314}, 2019.

\bibitem{mcdermott2019deep}
P.L. McDermott and C.K. Wikle.
\newblock Deep echo state networks with uncertainty quantification for
  spatio-temporal forecasting.
\newblock {\em Environmetrics}, 30(3):e2553, 2019.

\bibitem{lorenz1996predictability}
E.N. Lorenz.
\newblock Predictability: A problem partly solved.
\newblock In {\em Predcitibility of Weather and Climate}, volume~1, pages
  40--58, 1996.

\bibitem{pathak2017using}
J.~Pathak, Z.~Lu, B.R. Hunt, M.~Girvan, and E.~Ott.
\newblock Using machine learning to replicate chaotic attractors and calculate
  {L}yapunov exponents from data.
\newblock {\em Chaos}, 27(12):121102, 2017.

\bibitem{lu2017reservoir}
Z.~Lu, J.~Pathak, B.~Hunt, M.~Girvan, R.~Brockett, and E.~Ott.
\newblock Reservoir observers: Model-free inference of unmeasured variables in
  chaotic systems.
\newblock {\em Chaos}, 27(4):041102, 2017.

\bibitem{khodkar2019reduced}
M.A. Khodkar, P.~Hassanzadeh, S.~Nabi, and P.~Grover.
\newblock Reduced-order modeling of fully turbulent buoyancy-driven flows using
  the {G}reen's function method.
\newblock {\em Physical Review Fluids}, 4(1):013801, 2019.

\bibitem{jaeger2004harnessing}
H.~Jaeger and H.~Haas.
\newblock Harnessing nonlinearity: Predicting chaotic systems and saving energy
  in wireless communication.
\newblock {\em Science}, 304(5667):78--80, 2004.

\bibitem{Jaeger2007}
H.~Jaeger.
\newblock {E}cho state network.
\newblock {\em Scholarpedia}, 2(9):2330, 2007.
\newblock revision \#188245.

\bibitem{pascanu2013difficulty}
R.~Pascanu, T.~Mikolov, and Y.~Bengio.
\newblock On the difficulty of training recurrent neural networks.
\newblock In {\em International Conference on Machine Learning}, pages
  1310--1318, 2013.

\bibitem{goodfellow2016deep}
I.~Goodfellow, Y.~Bengio, and A.~Courville.
\newblock {\em Deep learning}.
\newblock MIT press, 2016.

\bibitem{lukovsevivcius2009reservoir}
M.~Luko{\v{s}}evi{\v{c}}ius and H.~Jaeger.
\newblock Reservoir computing approaches to recurrent neural network training.
\newblock {\em Computer Science Review}, 3(3):127--149, 2009.

\bibitem{gauthier2018reservoir}
D.J. Gauthier.
\newblock Reservoir computing: Harnessing a universal dynamical system.
\newblock {\em SIAM News}, 51:12, 2018.

\bibitem{pathak2018hybrid}
J.~Pathak, A.~Wikner, R.~Fussell, S.~Chandra, B.R. Hunt, M.~Girvan, and E.~Ott.
\newblock Hybrid forecasting of chaotic processes: Using machine learning in
  conjunction with a knowledge-based model.
\newblock {\em Chaos}, 28(4):041101, 2018.

\bibitem{zimmermann2018observing}
R.S. Zimmermann and U.~Parlitz.
\newblock Observing spatio-temporal dynamics of excitable media using reservoir
  computing.
\newblock {\em Chaos}, 28(4):043118, 2018.

\bibitem{lu2018attractor}
Z.~Lu, B.R. Hunt, and E.~Ott.
\newblock Attractor reconstruction by machine learning.
\newblock {\em Chaos}, 28(6):061104, 2018.

\bibitem{hochreiter1997long}
S.~Hochreiter and J.~Schmidhuber.
\newblock Long short-term memory.
\newblock {\em Neural Computation}, 9(8):1735--1780, 1997.

\bibitem{chen2015lstm}
K.~Chen, Y.~Zhou, and F.~Dai.
\newblock A {LSTM}-based method for stock returns prediction: A case study of
  {C}hina stock market.
\newblock In {\em 2015 IEEE International Conference on Big Data}, pages
  2823--2824. IEEE, 2015.

\bibitem{carbonneau2008application}
R.~Carbonneau, K.~Laframboise, and R.~Vahidov.
\newblock Application of machine learning techniques for supply chain demand
  forecasting.
\newblock {\em European Journal of Operational Research}, 184(3):1140--1154,
  2008.

\bibitem{cho2014learning}
K.~Cho, Bart V.M, C.~Gulcehre, D.~Bahdanau, F.~Bougares, H.~Schwenk, and
  Y.~Bengio.
\newblock Learning phrase representations using {RNN} encoder-decoder for
  statistical machine translation.
\newblock {\em arXiv preprint arXiv:1406.1078}, 2014.

\bibitem{graves2013speech}
A.~Graves, A.~Mohamed, and G.~Hinton.
\newblock Speech recognition with deep recurrent neural networks.
\newblock In {\em 2013 IEEE international Conference on Acoustics, Speech and
  Signal Processing}, pages 6645--6649. IEEE, 2013.

\bibitem{kim1999nonlinear}
H.S. Kim, R.~Eykholt, and J.D. Salas.
\newblock Nonlinear dynamics, delay times, and embedding windows.
\newblock {\em Physica D: Nonlinear Phenomena}, 127(1-2):48--60, 1999.

\bibitem{epanechnikov1969non}
V.A. Epanechnikov.
\newblock Non-parametric estimation of a multivariate probability density.
\newblock {\em Theory of Probability \& Its Applications}, 14(1):153--158,
  1969.

\bibitem{siegelmann1992computational}
H.T. Siegelmann and E.D. Sontag.
\newblock On the computational power of neural nets.
\newblock In {\em Proceedings of the fifth annual workshop on Computational
  learning theory}, pages 440--449. ACM, 1992.

\bibitem{yildiz2012re}
I.B. Yildiz, H.~Jaeger, and S.J. Kiebel.
\newblock Re-visiting the echo state property.
\newblock {\em Neural Networks}, 35:1--9, 2012.

\bibitem{mezic2005spectral}
I.~Mezi{\'c}.
\newblock Spectral properties of dynamical systems, model reduction and
  decompositions.
\newblock {\em Nonlinear Dynamics}, 41(1-3):309--325, 2005.

\bibitem{tu2014dynamic}
J.H. Tu, C.W. Rowley, D.M. Luchtenburg, S.L. Brunton, and J.N. Kutz.
\newblock On dynamic mode decomposition: Theory and applications.
\newblock {\em Journal of Computational Dynamics}, 1(2):391--421, 2014.

\bibitem{williams2015data}
M.O. Williams, I.G. Kevrekidis, and C.W. Rowley.
\newblock A data--driven approximation of the {K}oopman operator: Extending
  dynamic mode decomposition.
\newblock {\em Journal of Nonlinear Science}, 25(6):1307--1346, 2015.

\bibitem{arbabi2017ergodic}
H.~Arbabi and I.~Mezic.
\newblock Ergodic theory, dynamic mode decomposition, and computation of
  spectral properties of the {K}oopman operator.
\newblock {\em SIAM Journal on Applied Dynamical Systems}, 16(4):2096--2126,
  2017.

\bibitem{giannakis2017spatiotemporal}
D.~Giannakis, A.~Ourmazd, J.~Slawinska, and Z.~Zhao.
\newblock Spatiotemporal pattern extraction by spectral analysis of
  vector-valued observables.
\newblock {\em arXiv preprint arXiv:1711.02798}, 2017.

\bibitem{khodkar2018data}
M.A. Khodkar and P.~Hassanzadeh.
\newblock Data-driven reduced modelling of turbulent {Rayleigh--B}{\'e}nard
  convection using {DMD}-enhanced fluctuation--dissipation theorem.
\newblock {\em Journal of Fluid Mechanics}, 852, 2018.

\bibitem{sutskever2014sequence}
I.~Sutskever, O.l Vinyals, and Q.V. Le.
\newblock Sequence to sequence learning with neural networks.
\newblock In {\em Advances in Neural Information Processing Systems}, pages
  3104--3112, 2014.

\bibitem{yu2017long}
R.~Yu, S.~Zheng, A.~Anandkumar, and Y.~Yue.
\newblock Long-term forecasting using tensor-train {RNNs}.
\newblock {\em arXiv preprint arXiv:1711.00073}, 2017.

\bibitem{xingjian2015convolutional}
S.H.I. Xingjian, Z.~Chen, H.~Wang, D.~Yeung, W.~Wong, and W.~Woo.
\newblock Convolutional {LSTM} network: A machine learning approach for
  precipitation nowcasting.
\newblock In {\em Advances in Neural Information Processing Systems}, pages
  802--810, 2015.

\bibitem{ma2017walking}
Q.~Ma, L.~Shen, E.~Chen, S.~Tian, J.~Wang, and G.W. Cottrell.
\newblock Walking walking walking: Action recognition from action echoes.
\newblock In {\em Internationa Joint Conference on Artificial Intelligence},
  pages 2457--2463, 2017.

\bibitem{chantry2019scale}
M.~Chantry, T.~Thornes, T.~Palmer, and P.~D{\"u}ben.
\newblock Scale-selective precision for weather and climate forecasting.
\newblock {\em Monthly Weather Review}, 147(2):645--655, 2019.

\bibitem{bishop2006pattern}
C.M. Bishop.
\newblock {\em Pattern recognition and machine learning}.
\newblock springer, 2006.

\bibitem{kingma2014adam}
D.P. Kingma and J.~Ba.
\newblock Adam: A method for stochastic optimization.
\newblock {\em arXiv preprint arXiv:1412.6980}, 2014.

\bibitem{meng2019convergence}
Qi~Meng, Wei Chen, Yue Wang, Zhi-Ming Ma, and Tie-Yan Liu.
\newblock Convergence analysis of distributed stochastic gradient descent with
  shuffling.
\newblock {\em Neurocomputing}, 337:46--57, 2019.

\end{thebibliography}

\end{document}